\documentclass[letterpaper]{article} 
\usepackage{aaai24}  
\usepackage{times}  
\usepackage{helvet}  
\usepackage{courier}  
\usepackage[hyphens]{url}  
\usepackage{graphicx} 
\usepackage{natbib}  
\usepackage{caption} 
\usepackage{algorithm}
\usepackage{algorithmic}
\usepackage{times}
\usepackage{epsfig}
\usepackage{graphicx}
\usepackage{amsmath}
\usepackage{amssymb}
\usepackage{color}
\usepackage{multirow}
\usepackage{kotex}
\usepackage{colortbl}
\usepackage{xcolor}
\usepackage{oplotsymbl}
\usepackage{MnSymbol}

\usepackage{newfloat}
\usepackage{listings}
\DeclareCaptionStyle{ruled}{labelfont=normalfont,labelsep=colon,strut=off} 
\floatstyle{ruled}
\newfloat{listing}{tb}{lst}{}
\floatname{listing}{Listing}
\title{Open Domain Generalization with a Single Network\\
by Regularization Exploiting Pre-trained Features}

\author{
  Inseop Chung\textsuperscript{1}
  \,\,
  KiYoon Yoo\textsuperscript{1}
  \,\,
  Nojun Kwak\textsuperscript{1}
  \\
  \small{\texttt{\{jis3613, 961230, nojunk\}@snu.ac.kr}\\}
  \textsuperscript{1}Department of Intelligence and Information, \\ Graduate School of Convergence Science and Technology, \\ Seoul National University, Korea\\
}
\nocopyright

\begin{document}

\maketitle

\begin{abstract}
Open Domain Generalization (ODG) is a challenging task as it not only deals with distribution shifts but also category shifts between the source and target datasets. To handle this task, the model has to learn a generalizable representation that can be applied to unseen domains while also identify unknown classes that were not present during training. Previous work has used multiple source-specific networks, which involve a high computation cost. Therefore, this paper proposes a method that can handle ODG using only a single network. The proposed method utilizes a head that is pre-trained by linear-probing and employs two regularization terms, each targeting the regularization of feature extractor and the classification head, respectively. The two regularization terms fully utilize the pre-trained features and collaborate to modify the head of the model without excessively altering the feature extractor. This ensures a smoother softmax output and prevents the model from being biased towards the source domains. The proposed method shows improved adaptability to unseen domains and increased capability to detect unseen classes as well. Extensive experiments show that our method achieves competitive performance in several benchmarks. We also justify our method with careful analysis of the effect on the logits, features, and the head.
\end{abstract}

\section{Introduction}

Although deep neural networks have made remarkable achievements in the field of computer vision and its application, they often fail to generalize to out-of-distribution (OOD) data which are not seen during training due to the underlying assumption that the train and test data are independent and identically distributed. This is highly unlikely in real-world scenarios where the target samples at test time may have a discrepant distribution from the train set and may even change over time. Domain generalization (DG) addresses this issue by learning a generalizable representation with multiple source domains, but it assumes both source and target data share the same label set. In practice, the target domain may contain classes that do not exist in the source domains. Furthermore, the label sets may differ even among the source domains. Recently, \cite{shu2021open} proposed a very challenging problem called \textbf{Open Domain Generalization (ODG)} which assumes that the target domain has not only a different data distribution but also a different label set from the source domains and each source domain holds a different label set as well.

\begin{figure}[t]
    \centering
    \includegraphics[width = 0.9\linewidth]{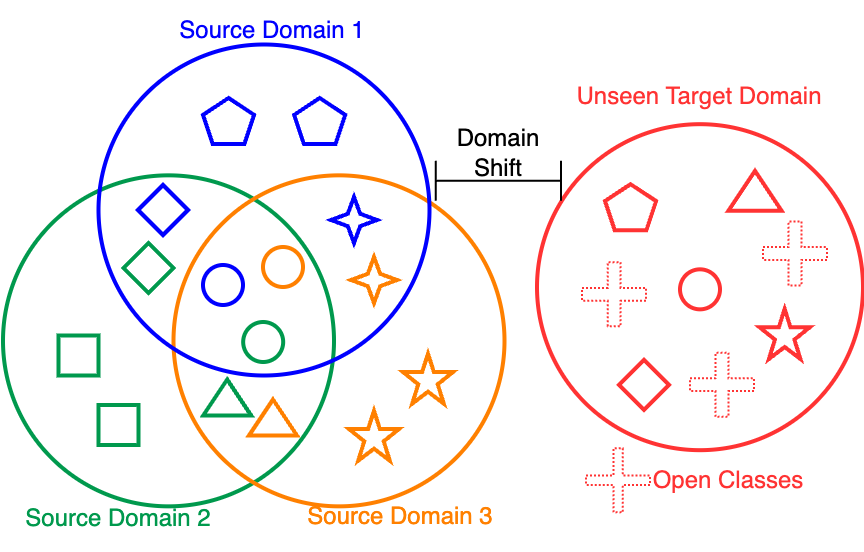}
    \caption{Concept of Open Domain Generalization. Each source domain has a different label set. The target domain has overlapping classes with the source domains as well as the open classes which do not exist in the source domains.}
    \label{fig:odg_concept}
\end{figure}

Fig.\ref{fig:odg_concept} shows the concept of ODG in which each source domain has a disparate label set and the target domain holds both the classes in the source domains (known classes) and that are not, which are called \emph{open} classes. 
ODG aims to achieve two objectives: firstly, to train a model that can generalize effectively to an unseen target domain that exhibits a significant distribution shift from the source domains, and secondly, to recognize the open classes that were not encountered during the training process. The red crosses with dotted lines in Fig.\ref{fig:odg_concept} indicate the open classes in the target domain. The open classes may consist of multiple classes but it is treated as a single \emph{open} class since the goal is to recognize it as an unknown class regardless of its actual class. What makes this problem more challenging is the disparity of label sets among the source domains: some classes exist in multiple domains while the minor classes only exist in a certain source domain (e.g. $\pentago$, $\square$, $\largestar$ in Fig.\ref{fig:odg_concept}). This challenging setting makes it difficult to apply the existing DG methods~\cite{li2018learning, li2018deep, muandet2013domain} which assume the source domains share the same label set. \\
\indent To address this problem, \cite{shu2021open} proposes a Domain-Augmented Meta-Learning (DAML) which employs meta-learning and domain augmentation methods to learn a representation generalizable to the unseen target domain. However, it requires a separate network for each source domain and uses the \emph{ensemble} of all the source domain networks as the final prediction for the target domain input. This is highly undesirable in real-world scenarios where a variety of source domains can be included in the training set. It implies that as the number of source domains increases, the number of required networks and the inference cost increase linearly as well. It can be technically infeasible under a practical environment with limited resources such as power, memory, and computational budget. \\

\begin{figure}[t]
    \centering
    \includegraphics[width = 0.9\linewidth]{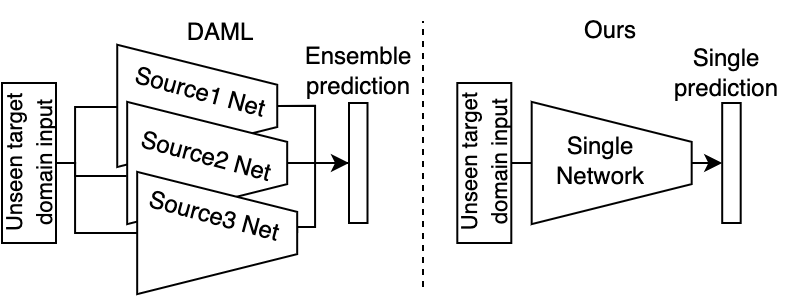}
    \caption[DAML vs Ours]{DAML uses an ensemble of multiple source domain networks while our proposed method uses only one network.}
    \label{fig:ours_daml}
\end{figure}

\indent Therefore, in this paper, we tackle ODG only with a single network (see Figure~\ref{fig:ours_daml}). Since domain generalization typically assumes access to a pre-trained network, our main motivation is to fully exploit this. To safeguard the feature extractor from fitting too much to the source domains, we first train only the classification head instead of fully fine-tuning all the model parameters following previous works \cite{kumar2022finetuning, levine2016end, kanavati2021partial} such that the readily trained classification head allows the feature extractor not to adapt too much during fine-tuning. 

To further guide the network during fine-tuning, we propose two regularization methods for the feature extractor and the head respectively. For the feature extractor, we make a prototype of each class in the source domains using the features of the \emph{pre-trained} feature extractor. Then, during fine-tuning, we minimize the distance between the feature of each input and its corresponding prototype. We also regulate the head by maximizing the entropy of its softmax output given the \emph{pre-trained} feature of each source input. 
The first term aims to promote the clustering of features by classes and minimize the distortion of pre-trained features. Meanwhile, the second term encourages the head to deviate from the pre-trained head. When the proposed terms are optimized along with the standard cross-entropy loss, they effectively adjust the head to a better solution while preserving the pre-trained features. We observe this results in a smoother class probability distribution (i.e., softmax output) which prevents the model from making overly confident predictions on the unknown classes and prohibiting it from being biased towards the source domains, allowing for better adaptability to the unseen domains.

\indent We verify that the proposed method makes our features deviate less from the pre-trained features and observe that the more the head changes from its pre-trained parameters, the more the performance gain. We empirically show that our proposed method boosts performance under the ODG setting in that it outperforms DAML in several benchmarks using only a single network. For open class recognition, as \cite{vaze2022openset} claims that a good closed-set (known classes) classifier is a good open-set classifier, we focus more on improving the accuracy of known classes to achieve better performance of open class recognition. Note that the purpose of our work is not to propose a SOTA method but to demonstrate that fully exploiting the pre-trained features can improve the performance in ODG with only a single network, greatly reducing the inference cost.

\section{Related Work}

\textbf{Domain Generalization} is a task of training a model on source domains to learn a representation that generalizes well to unseen target domains. The most popular solution is to learn a domain-invariant feature representation across source domains~\cite{ghifary2015domain, ghifary2016scatter, li2018domain, li2018deep, muandet2013domain}. Data augmentation is another strategy employed. \citet{shankar2018generalizing} directly perturbed the inputs by the gradients of loss to augment the inputs, while \citet{zhou2020learning} learned a generator model to diversify the source domains. \citet{zhou2021domain} mixed the style of training instances to generate novel domains. Meta-learning approach~\cite{li2018learning, dou2019domain} is also widely used to simulate domain shifts. \citet{saito2019semi} proposed a minimax entropy approach for semi-supervised domain adaptation. \citet{zhao2020domain} proposed an entropy-regularization approach to learn domain-invariant features. \citet{cite:ECCV20Cumix} introduced mixing up source samples during training to recognize unseen classes in unseen domains. \nocite{piratla2020efficient} \citet{kim2021selfreg} proposed a regularization method based on self-supervised contrastive learning. \citet{cha2021swad} theoretically showed that finding flat minima results in better generalization and proposed to sample weights densely for stochastic weights averaging~\cite{izmailov2018averaging}. \citet{cha2022domain} exploited an oracle model and maximized the mutual information between the oracle and the target model. \citet{shu2021open} primarily proposed an ODG method by combining DG and open class recognition. It utilized the meta-learning scheme along with domain augmentation. Recently, \citet{zhu2021crossmatch} proposed another ODG problem having only a single source domain, it used an adversarial data augmentation strategy to generate auxiliary samples and adopted a consistency regularization on them. \citet{kumar2022finetuning} did not particularly tackle the domain generalization but it showed that fine-tuning can distort the pre-trained features and underperform on OOD data. It insists that training the head prior to fine-tuning improves the performance on the OOD data. Our work tackles the same problem proposed in \cite{shu2021open} and competes with it, but our method uses only a single network unlike \cite{shu2021open} which utilizes multiple source domain networks.

\noindent \textbf{Open-Set Recognition} is a task of identifying whether an input belongs to the classes seen during training or not. The open classes are the classes that are not included in the training set but present at the test time. In the open-set setting, the model has to not only accurately predict a class in the closed-set but also distinguish samples belonging to the open classes as ``unknown". \citet{liang2017enhancing} claimed that using temperature scaling and adding small perturbations to the input can separate the softmax score between the known and open classes. \citet{zhou2021learning} proposed placeholders for both data and the classifier to transform closed-set training into open-set training. Recently, \citet{vaze2022openset} presented that there is a high correlation between the closed-set and open-set accuracies and showed that open-set performance can be enhanced by improving the closed-set accuracy. Based on this analysis, we resolve the open-set recognition by improving the performance on the closed-set of the target domain. 

\section{Methods}
\subsection{Problem Definition}

Open domain generalization assumes multiple source domains are available during training, $\{D^{i}\}^K_{i=1}$ where $K$ refers to the number of source domains. Each source domain consists of $N^i$ input-label pairs $D^i = \{(x^i_j, y^i_j)\}^{N^i}_{j=1}$ with its own label set, $C^i = \bigcup_{j} \{y_j^i\}$.
As illustrated in Fig.~\ref{fig:odg_concept}, some classes are shared across the source domains while others only belong to one source domain. $C = \bigcup_{i=1}^{K}C^{i}$ is the union of all the classes in the source domains. The goal is to train the model only with the source domain data and learn a representation generalizable to the unseen target domain $D^t$. $D^t$ contains all or subset of $C$ (known classes) and the open classes (unknown classes) which do not exist in $C$. At evaluation time, the source-trained model is given inputs from $D^t$ and has to correctly classify the input if it belongs to one of the classes in $C$, otherwise recognize it as the open class. Note that $D^t$ is only used at test time for evaluation and not used at all during training.
 
\subsection{Preliminary: LP-FT} 
\label{preliminary}
One of our main goals is to train the feature extractor only to an extent such that the pre-trained features remain favorable to the target domain. Having access only to the source domains, we can achieve this by preventing the feature extractor from overfitting to the source domains.
Earlier works in transfer learning have used partial fine-tuning or linear probing to mitigate catastrophic forgetting of the pre-trained knowledge \shortcite{howard2018universal, xie2021composed, zhai2022lit}. 
To realize our goal, we draw inspiration from a recent analysis \cite{kumar2022finetuning}, which shows that higher performance on OOD data\footnote{OOD data refers to the data with a different distribution from the in-distribution (ID) data (data seen during training). In domain generalization, the target domain can be considered as the OOD data.} can be achieved when linear-probing (LP: updating only the classifier head while freezing the feature extractor layers) is followed by fine-tuning (FT: updating all the model parameters) given a good pre-trained features. We refer to this training scheme as LP-FT following the original work.

Following \cite{shu2021open}, we employ a classification network consisting of a feature extractor $f$ and a head $h$. $f$ is initialized with pre-trained weights (we use ImageNet~\cite{deng2009imagenet} pre-trained weights.) and $h$ is a $C$-way linear classifier. We first train \emph{only the head} by linear-probing:

\begin{equation}\label{eq:lp}
\small
\mathcal{L}_{\text{lp}} = \sum^K_i \sum^{N^i}_j CE(h(\bar{f_0}(x^i_j)), y^i_j), 
\end{equation}

\noindent where $CE$ and $f_0$ refer to the cross-entropy loss and the pre-trained $f$ respectively. We denote the frozen $f_0$ by $\bar{f_0}$. We obtain $h^{lp}$, a linear classifier head trained on the pre-trained features, by minimizing $\mathcal{L}_{\text{lp}}$. We then initialize the head with $h^{lp}$ and update all model parameters of $f$ and $h$.

\begin{equation}\label{eq:lp-ft}
\small
\mathcal{L}_{\text{lp-ft}} = \sum^K_i \sum^{N^i}_j CE(h^{lp}(f(x^i_j)), y^i_j). \\
\end{equation}

\noindent In practice, we train the model with mini-batches, but we omit the details for simpler notations. The intuition of LP-FT is that with a randomly initialized head, FT distorts ID features more and changes OOD features less which makes the head overfit to the distorted ID features and leads to poor performance on OOD data. It resolves this problem by initializing with a pre-trained head via linear-probing which preserves the pre-trained features favorable for both ID and OOD data, reducing the amount of change in the feature extractor. Our method adopts the philosophy of LP-FT and further boosts the performance on OOD data, which is the target domain data in our setting.

\begin{figure}[t]
    \centering
    \includegraphics[width = 0.85\linewidth]{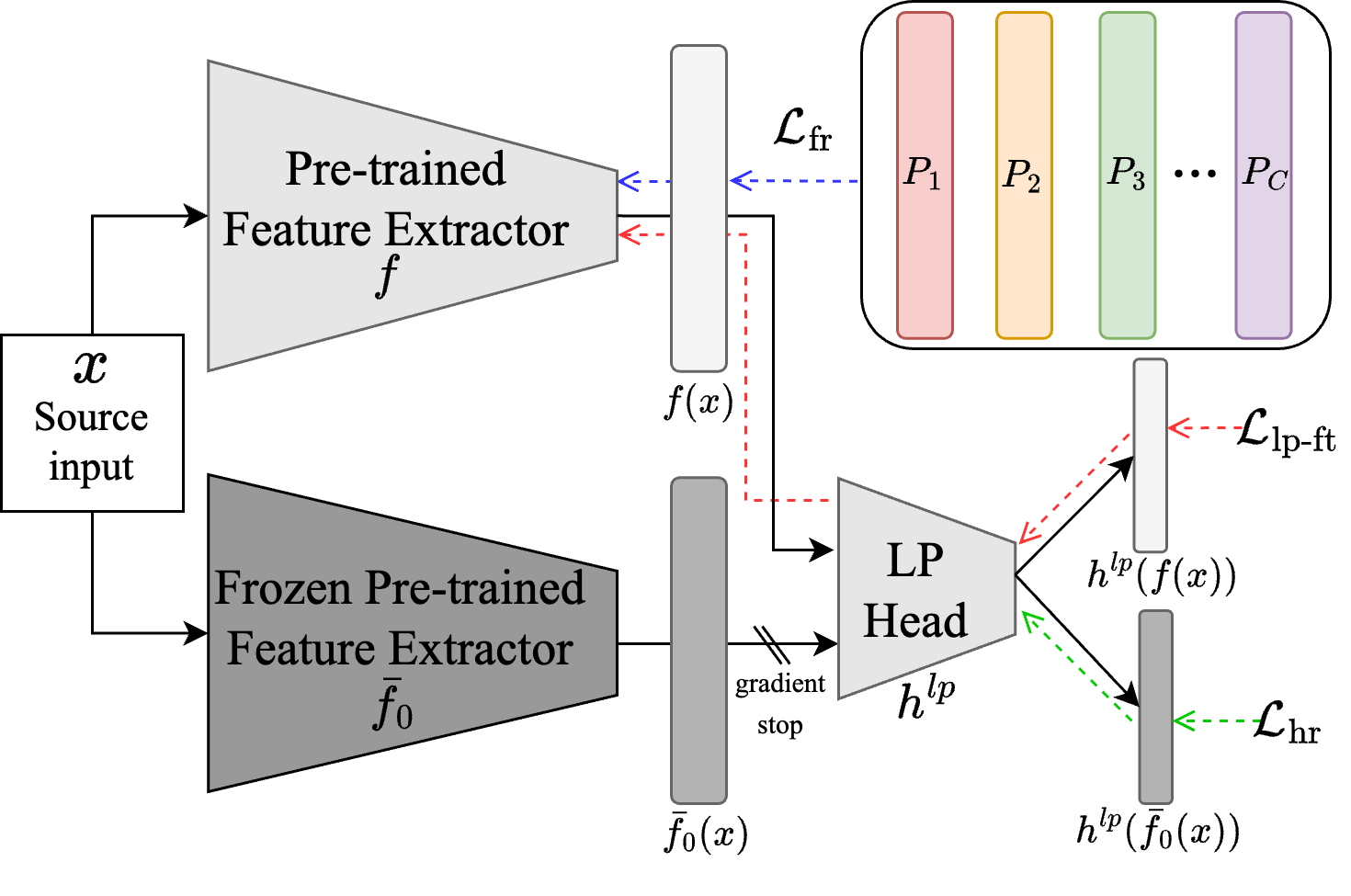}
    \caption{Overall schematic of our proposed training scheme. Each dotted line indicates the path of back-propagation for each loss. Note that $\mathcal{L}_{\text{hr}}$ back-propagates only upto the head and not to $f$. Also, no loss back-propagates to $\bar{f_0}$, which is only employed during training to provide the pre-trained features of the source domain inputs.
    }
    \label{fig:loss_concept}
\end{figure}

\subsection{Clustering Features by Pre-trained Prototype}
With the pre-trained head by linear-probing, we expect that clustering the source domain features according to class would also help the target domain features cluster together class-wisely.
We generate a prototype that works as a centroid for each class-wise cluster and minimize the distance between each source input feature and its corresponding class' prototype.
Since the goal of LP-FT is to not distort the pre-trained features too much, we choose to generate the prototype of each class by averaging the features from the pre-trained feature extractor $f_0$:

\begin{equation}\label{eq:proto}
\small
P_c = \frac{1}{\sum^K_i N^i_c} \sum^K_i \sum^{N^i_c}_j (\bar{f_0}(x^i_j)),
\end{equation}

\noindent where $c$ is the class index and $P_c$ is the generated prototype of class $c$. $N^i_c$ is the number of samples labeled as class $c$ in domain $i$. Therefore, $P_c$ can be a prototype across multiple source domains if the class $c$ belongs to more than one source domain. 
We then minimize the distance between each source input feature and its prototype as follows:

\begin{equation}\label{eq:fr}
\small
\mathcal{L}_{\text{fr}} = \sum^K_i \sum^C_c \sum^{N^i_c}_j ||f(x^i_j) - P_c||^2_2. \\
\end{equation}

\noindent Note that each $P_c$ is generated before the training and saved in memory. All $P_c$'s are pre-defined and \emph{fixed} during training, hence the gradient of $\mathcal{L}_{\text{fr}}$ only back-propagates to $f$, the target feature extractor that we want to train, but not to $P_c$. Along with $\mathcal{L}_{\text{lp-ft}}$, $\mathcal{L}_{\text{fr}}$ regulates $f$ to cluster the features  centered around the prototype of the class the inputs belong to and prevents distorting the pre-trained features excessively. Please be informed that $f$ in $\mathcal{L}_{\text{lp-ft}}$ and $\mathcal{L}_{\text{fr}}$ are initialized as $f_0$ and its parameters are updated as the training proceeds.

\subsection{Maximizing Entropy of Pre-trained Features}
The head regularization term maximizes the entropy of the classifier's softmax output conditioned on the pre-trained features of the source domain inputs. We employ a frozen pre-trained feature extractor $\bar{f_0}$ which is a separate feature extractor different from the target $f$ we want to train by $\mathcal{L}_{\text{lp-ft}}$ and $\mathcal{L}_{\text{fr}}$. With the head initialized as $h^{lp}$, we produce the logit for each pre-trained source feature as follows: $\sigma(z^s) = \text{softmax}(h^{lp}(\bar{f_0}(x^s))) \in \mathbb{R}^C$ where $z=h^{lp}(\bar{f_0}(x))$ is the logit (the raw output of the head) and $\sigma$ is the softmax operation. We minimize the following loss term to \emph{maximize} its entropy: 

\begin{equation}\label{eq:hr}
\small
\mathcal{L}_{\text{hr}} = \sum^K_i \sum^{N^i}_j \sum^{C}_c \sigma(z^i_j)^c\cdot\log\sigma(z^i_j)^c, \\
\end{equation}

\noindent where $\sigma(z)^c$ denotes the $c$-th element of the softmax output $\sigma(z)$, or the probability of class $c$.
Note that the loss term \emph{only updates the parameters of the head $h$, and does not affect the feature extractor, $f$}.

\begin{figure}[t]
    \centering
    \includegraphics[width = 1.0\linewidth]{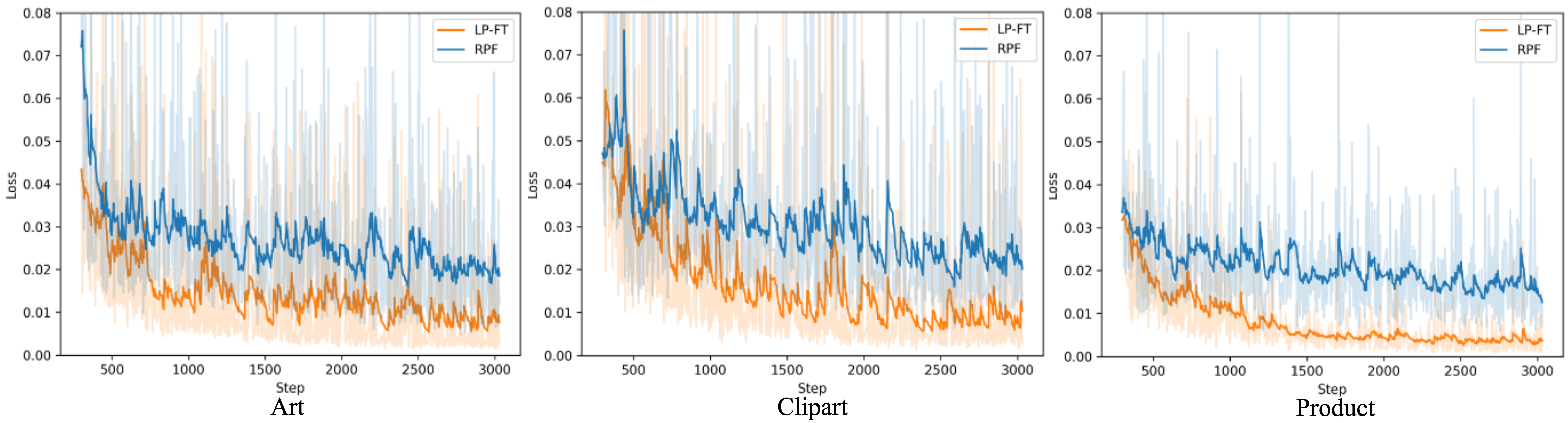}
    \caption{ The plot of $\mathcal{L}_{\text{lp-ft}}$ for each source domain in the Office-Home dataset when the target domain is Real-World. The loss decreases for both LP-FT and RPF, but RPF clearly shows a higher loss scale. Results on other target domains are in the appendix.
    }
    \label{fig:train_loss}
\end{figure}

\subsection{Overall Objective}
The overall loss function of our proposed method is as follows:

\begin{equation}\label{eq:overall}
\small
\mathcal{L}_{\text{total}} = \mathcal{L}_{\text{lp-ft}} + \mathcal{L}_{\text{fr}} + \lambda_{hr}\mathcal{L}_{\text{hr}}.  \\
\end{equation}

\noindent Fig.~\ref{fig:loss_concept} shows the overall training scheme of our proposed method. The three loss terms are summed and optimized simultaneously. $\lambda_{hr}$ is the balance weight for $\mathcal{L}_{\text{hr}}$. $\mathcal{L}_{\text{lp-ft}}$ affects both $f$ and $h$, while $\mathcal{L}_{\text{fr}}$ and $\mathcal{L}_{\text{hr}}$ affect only $f$ and $h$ respectively. 
$\mathcal{L}_{\text{fr}}$ regulates $f$ to cluster the features by class and not to deviate too far from the pre-trained features by anchoring each feature to its corresponding prototype, $P_c$. 
On the other hand, $\mathcal{L}_{\text{hr}}$ regulates the head to maximize the entropy of pre-trained features of source inputs from $f_0$. Since the head is initialized as $h^{lp}$ which is fitted to classify the pre-trained features from $f_0$ correctly, minimizing $\mathcal{L}_{\text{hr}}$ demands the head to deviate from $h^{lp}$  to predict less confidently (higher entropy) on the pre-trained features. The combination of these two terms results in a larger $\mathcal{L}_{\text{lp-ft}}$ as shown in Fig.\ref{fig:train_loss}.
Larger $\mathcal{L}_{\text{lp-ft}}$ indicates that the softmax output has smoother class probability distribution. It implies that the model is less biased towards the source domains seen during training. and less prone to making over-confident predictions on the unknown classes which leads to improved adaptability of the model to the unseen domains containing unknown classes.

However, a high weight on $\mathcal{L}_{\text{hr}}$ can impair the head, thus we use a small $\lambda_{hr}$. We empirically find that $\lambda_{hr}=0.1$ works reasonably well across the benchmarks. Based on our empirical observations, our proposed regularization terms have the effect of smoothing the final output and inducing greater changes to the head parameters relative to its initial weight $h^{lp}$. We have also discovered that our method helps to reduce the domain gap between the source and target domains, encourages greater clustering of target features (thus minimizing intra-class variance), and results in less deviation from the pre-trained features. We provide a more detailed analysis of these findings in experiments. We name our method (training by $\mathcal{L}_{\text{total}}$) RPF which stands for Regularization through Pre-trained Features.

\subsection{Inference}
At inference time, the model is given target domain inputs that are never seen during training. We only use a single network trained by $\mathcal{L}_{\text{total}}$ as opposed to DAML which uses an \textbf{ensemble} of multiple networks. 
When three source domains are used, DAML requires three separate networks and it takes 10.91 GFLOPs and 8.37 milliseconds for an inference of a single image while our proposed method only takes 3.64 GFLOPs and 3.25 milliseconds on a single NVIDIA RTX 3090.
Note that $\bar{f_0}$ used in (\ref{eq:hr}) is \emph{\textbf{not}} needed at inference time; it is employed only at training time to perform head regularization. Following ~\cite{you2019universal}, we set a threshold on the confidence score and classify the target input as the open class if its maximum confidence score ($\max\limits_{c\in C} \sigma(z^t)^c$) is lower than the threshold. Otherwise, it is classified as the class of the maximum confidence score.

\begin{table*}[t]
        \small
	\begin{center}
		\resizebox{0.95\textwidth}{!}{
				\begin{tabular}{ll|cccccccc|cc}
					\hline
					& &\multicolumn{2}{c}{\textbf{Real-World}}&\multicolumn{2}{c}{\textbf{Product}}&\multicolumn{2}{c}{\textbf{Clipart}}&\multicolumn{2}{c|}{\textbf{Art}}&\multicolumn{2}{c}{\textbf{Avg}} \\
					\textbf{Model} &\textbf{Method}&Acc&H-score&Acc&H-score&Acc&H-score&Acc&H-score&Acc&H-score \\
					\hline
					I & w/o HR and FR (LP-FT) & $68.44$ & $61.74$ & $60.53$ & $57.04$ & $43.56$ & $42.24$ & $53.07$ & $50.07$ & $56.40\pm{0.42}$ & $52.77\pm{0.76}$ \\
					II & w/o HR & $68.77$ & $61.65$ & $61.62$ & $58.39$ & $44.24$ & $42.33$ & $54.50$ & $52.03$ & $57.28\pm{0.88}$ & $53.60\pm{0.59}$ \\
                    III & w/o FR & $68.90$ & $61.43$ & $61.48$ & $57.61$ & $44.33$ & $43.44$ & $53.67$ & $48.56$ & $57.09\pm{0.59}$ & $52.76\pm{0.33}$ \\
                    IV & w/o pre-trained head & $65.08$ & $58.28$ & $56.78$ & $55.07$ & $41.08$ & $41.20$ & $52.42$ & $48.28$ & $53.84\pm{0.47}$ & $50.71\pm{0.62}$ \\
                    V & \emph{HR$_{f}$} & $67.93$ & $60.94$ & $61.03$ & $58.11$ & $43.47$ & $42.31$ & $55.34$ & $52.16$ & $56.94\pm{0.50}$ & $53.38\pm{0.40}$ \\
                    VI & \emph{Ent-Min HR} & $67.28$ & $60.07$ & $61.06$ & $57.47$ & $44.02$ & $42.22$ & $54.92$ & $52.05$ & $56.82\pm{0.33}$ & $52.95\pm{0.61}$ \\
                    \hline
                    VII & RPF & $\mathbf{70.67}$ & $\mathbf{61.92}$ & $\mathbf{63.51}$ & $\mathbf{59.55}$ & $\mathbf{44.55}$ & $\mathbf{43.55}$ & $\mathbf{56.05}$ & $\mathbf{53.02}$ & $\mathbf{58.70}\pm{0.46}$ & $\mathbf{54.51}\pm{0.57}$ \\
					\hline
				\end{tabular}%
		}
	\end{center}
        \caption{Ablation study on the Office-Home dataset in the open-domain setting. \emph{HR$_{f}$} and \emph{Ent-Min HR} are trained with FR.}
        \label{table:abl}
\end{table*}

\begin{table}[t]
        \small
	\begin{center}
		\resizebox{0.9\linewidth}{!}{
				\begin{tabular}{l|c|c|cccc|c}
					\hline
					\textbf{Metric} & \textbf{Domain(s)} & \textbf{Method} & \textbf{R} & \textbf{P} & \textbf{C} & \textbf{A} & \textbf{Avg} \\
					\hline
					\multirow{4}{*}{Domain gap} 
                        & \multirow{2}{*}{Trg. vs. Src} 
                            & RPF & $0.16$ & $0.16$ & $0.27$ & $0.15$ & $\mathbf{0.19}$ \\ 
                            & & LP-FT & $0.19$ & $0.26$ & $0.30$ & $0.22$ & $0.24$ \\
                          \cline{2-8}
                        & \multirow{2}{*}{Source}  
                        & RPF & $0.30$ & $0.20$ & $0.30$ & $0.19$ & $\mathbf{0.25}$ \\
                        &  & LP-FT & $0.37$ & $0.33$ & $0.33$ & $0.29$ & $0.33$ \\
                    \hline
                    \multirow{4}{*}{Intra-class dist.} 
                        & \multirow{2}{*}{Target} 
                            & RPF & $0.44$ & $0.30$ & $0.49$ & $0.36$ & $\mathbf{0.40}$ \\
                            & & LP-FT  & $0.51$ & $0.48$ & $0.55$ & $0.58$ & $0.53$ \\
                        \cline{2-8}
                        & \multirow{2}{*}{Source}  
                              & RPF  & $0.35$ & $0.26$ & $0.40$ & $0.28$ & $\mathbf{0.32}$ \\
                              & & LP-FT  & $0.43$ & $0.44$ & $0.44$ & $0.44$ & $0.44$ \\
                    \hline
                    \multirow{4}{*}{Diff. from $f_0$} 
                        & \multirow{2}{*}{Target}  
                        & RPF & $0.59$ & $0.61$ & $0.66$ & $0.46$ & $\mathbf{0.58}$ \\
                        & & LP-FT  & $0.64$ & $0.71$ & $0.69$ & $0.54$ & $0.64$ \\
                    \cline{2-8}
                    & \multirow{2}{*}{Source}  
                      & RPF  & $0.65$ & $0.57$ & $0.72$ & $0.65$ & $\mathbf{0.65}$ \\
                      & & LP-FT  & $0.73$ & $0.69$ & $0.73$ & $0.75$ & $0.73$ \\
					\hline
				\end{tabular}%
		}
	\end{center}
        \caption{Analysis on domain gap, intra-class distance and difference from pre-trained features.}
        \label{table:f_anal}
\end{table}

\begin{table}[t]
        \small
	\begin{center}
		\resizebox{0.9\linewidth}{!}{
				\begin{tabular}{l|c|c|cccc|c}
					\hline
					\textbf{Metric} & \textbf{Class} &\textbf{Method} & \textbf{R} & \textbf{P} & \textbf{C} & \textbf{A} & \textbf{Avg} \\
					\hline
					\multirow{4}{*}{Confidence score} & 
                        \multirow{2}{*}{Known} 
                            & RPF  & $0.76$ & $0.72$ & $0.57$ & $0.65$ & $\mathbf{0.67}$ \\ 
                              & & LP-FT & $0.81$ & $0.77$ & $0.59$ & $0.70$ & $0.72$ \\
                      \cline{2-8}
                        & \multirow{2}{*}{Unknown}
                          & RPF  & $0.56$ & $0.49$ & $0.47$ & $0.50$ & $\mathbf{0.51}$ \\
                          & & LP-FT  & $0.63$ & $0.58$ & $0.50$ & $0.59$ & $0.57$ \\
                    \hline
                    \multirow{4}{*}{Entropy} & 
                        \multirow{2}{*}{Known} & 
                        RPF  & $0.83$ & $1.04$ & $1.56$ & $1.24$ & $\mathbf{1.17}$ \\ 
                          & & LP-FT  & $0.65$ & $0.80$ & $1.46$ & $1.01$ & $0.98$ \\
                          \cline{2-8}
                        & \multirow{2}{*}{Unknown} & 
                          RPF & $1.53$ & $1.80$ & $1.96$ & $1.78$ & $\mathbf{1.77}$ \\
                          & & LP-FT & $1.23$ & $1.42$ & $1.84$ & $1.40$ & $1.45$ \\
					\hline
				\end{tabular}%
		}
	\end{center}
        \caption{Analysis on maximum confidence score and entropy of logits using the target domain inputs.}
        \label{table:l_anal}
\end{table}

\begin{table}[t]
        \small
	\begin{center}
		\resizebox{0.9\linewidth}{!}{
				\begin{tabular}{l|c|cccc|c}
					\hline
					\textbf{Metric} & \textbf{Method} & \textbf{R} & \textbf{P} & \textbf{C} & \textbf{A} & \textbf{Avg} \\
					\hline
					\multirow{2}{*}{Euclidean Dist.} & RPF & $0.1072$ & $0.1412$ & $0.0848$ & $0.1416$ & $0.1187$ \\ 
                          & LP-FT & $0.1022$ & $0.0852$ & $0.1104$ & $0.1129$ & $0.1027$ \\
                    \hline 
                    \multirow{3}{*}{Imp.1(\%)} & RPF & $6.87$ & $4.17$ & $12.69$ & $8.10$ & $7.96$ \\ 
                          & LP-FT & $4.25$ & $-0.80$ & $11.13$ & $2.55$ & $4.28$ \\ 
                          & $\Delta$ & $2.62$ & $4.98$ & $1.56$ & $5.56$ & $3.68$ \\ 
                    \hline
                    \multirow{3}{*}{Imp.2(\%)} & RPF & $4.26$ & $-9.98$ & $1.44$ & $-6.93$ & $-2.81$ \\ 
                          & LP-FT & $6.17$ & $1.95$ & $4.59$ & $3.77$ & $4.12$ \\ 
                          & $\Delta$ & $-1.91$ & $-11.93$ & $-3.15$ & $-10.70$ & $-6.92$  \\
                    \hline
				\end{tabular}%
		}
	\end{center}
        \caption{Head parameters similarity and performance improvement over the linear-probing model.}
        \label{table:lp_compare}
\end{table}

\section{Experiments}
\subsection{Experimental settings}
In domain generalization, a dataset consists of multiple domains and shares the same label set (classes) across the domains. The experiment is performed by changing the target domain and using the remaining domains as the source domains. However, in ODG proposed in \cite{shu2021open}, the label set is different between domains as depicted in Fig.\ref{fig:odg_concept}. \citet{shu2021open} provides the class split between the domains and other detailed information needed to implement their ODG setting. We exactly follow these experimental settings for a fair comparison.
Experiments are conducted under three different benchmarks, \textbf{PACS}~\cite{cite:ICCV17DBA}, \textbf{Office-Home}~\cite{cite:CVPR17OfficeHome} and \textbf{Multi-Datasets} scenario. \textbf{Multi-Datasets} scenario is a practical setting proposed by \citet{shu2021open} where the model is trained on three public datasets, \textbf{Office-31}~\cite{cite:ECCV10Office}, \textbf{STL-10}~\cite{coates2011analysis} and \textbf{Visda2017}~\cite{visda2017}, and evaluated on four different domains of \textbf{Domain-Net}~\cite{peng2019moment} which is another benchmark for cross-domain generalization. Its purpose is to simulate the situation where the source domains are obtained from different resources. 
Experiments are conducted using a ResNet-18~\cite{he2016deep} backbone pre-trained on ImageNet~\cite{deng2009imagenet} with a head whose output dimension is defined as the number of classes in the source domains. Detailed information about datasets, class splits and implementation details are explained in the appendix.
We report the average of 3 runs for each experiment. Two metrics are used for the evaluation: 1) Acc, which is the accuracy only on the known classes (open classes are not given when calculating Acc) and 2) H-score, which is the harmonic mean of known and open class accuracy (both known and open classes are given) following ~\cite{Fu_2020_ECCV, shu2021open}.

\begin{table*}[t]
        \small
	\begin{center}
		\resizebox{1.0\textwidth}{!}{
				\begin{tabular}{ll|cccccccc|cc}
					\hline
					& &\multicolumn{2}{c}{\textbf{Art}}&\multicolumn{2}{c}{\textbf{Sketch}}&\multicolumn{2}{c}{\textbf{Photo}}&\multicolumn{2}{c|}{\textbf{Cartoon}}&\multicolumn{2}{c}{\textbf{Avg}} \\
					& \textbf{Method}&Acc&H-score&Acc&H-score&Acc&H-score&Acc&H-score&Acc&H-score \\
					\hline
                    \multirow{6}{*}{\small Single}
                    & FC~\cite{li2019feature} &$51.12$&$39.01$&$51.15$&$\cellcolor{yellow!25}49.28$&$60.94$&$45.79$&$\cellcolor{yellow!25}69.32$&$52.67$&$58.13\pm{0.20}$&$46.69\pm{0.25}$\\
                    & PAR~\cite{cite:NIPS19PAR} &\cellcolor{yellow!25} $52.97$&$39.21$&$53.62$&$\cellcolor{orange!25}52.00$&$51.86$&$36.53$&$67.77$&$52.05$&$56.56\pm{0.51}$&$44.95\pm{0.57}$\\
                    & RSC~\cite{cite:ECCV20RSC} &$50.47$&$38.43$&$50.17$&$44.59$&$67.53$&$49.82$&$67.51$&$47.35$&$58.92\pm{0.46}$&$45.05\pm{0.60}$\\
                    & CuMix~\cite{cite:ECCV20Cumix}&\cellcolor{orange!25} $53.85$&$38.67$&$37.70$&$28.71$&$65.67$&$49.28$&$\cellcolor{red!25}\mathbf{74.16}$&$47.53$&$57.85\pm{0.32}$&$41.05\pm{0.66}$\\
                    & LP-FT~\cite{kumar2022finetuning} & $49.12$ & $\cellcolor{orange!25}44.27$ & $\cellcolor{yellow!25}55.25$ & $43.50$ & $\cellcolor{orange!25}69.50$ & $\cellcolor{orange!25}67.75$ & $62.73$ & $\cellcolor{orange!25}57.94$ & $\cellcolor{yellow!25}59.15\pm{0.83}$ & $\cellcolor{orange!25}53.36\pm{0.90}$ \\
                    & RPF (Ours) & $50.99$ & $\cellcolor{red!25}45.87$ & $\cellcolor{orange!25}55.55$ & $49.00$ & $\cellcolor{yellow!25}69.33$ & $\cellcolor{red!25}\mathbf{70.51}$ & $67.87$ & $\cellcolor{red!25}59.47$ & $\cellcolor{orange!25}60.94\pm{0.78}$ & $\cellcolor{red!25}\mathbf{56.21}\pm{0.64}$ \\
					\hline
                    \multirow{1}{*}{\small Ensemble} 
                    & DAML~\cite{shu2021open} &\cellcolor{red!25}$\mathbf{54.10}$&$\cellcolor{yellow!25}43.02$&$\cellcolor{red!25}58.50$&$\cellcolor{red!25}\mathbf{56.73}$&$\cellcolor{red!25}\mathbf{75.69}$&$\cellcolor{yellow!25}53.29$&$\cellcolor{orange!25}73.65$&$\cellcolor{yellow!25}54.47$&\cellcolor{red!25}$\mathbf{65.49}\pm{0.36}$&$\cellcolor{yellow!25}51.88\pm{0.42}$\\
                    \hline
				\end{tabular}%
		}
	\end{center}
	\caption{Results on PACS dataset in the open-domain setting.}
        \label{table:pacs}
\end{table*}

\begin{table*}[t]
        \small
	\begin{center}
		\resizebox{1.0\textwidth}{!}{
				\begin{tabular}{ll|cccccccc|cc}
					\hline
					& &\multicolumn{2}{c}{\textbf{Real-World}}&\multicolumn{2}{c}{\textbf{Product}}&\multicolumn{2}{c}{\textbf{Clipart}}&\multicolumn{2}{c|}{\textbf{Art}}&\multicolumn{2}{c}{\textbf{Avg}} \\
					& \textbf{Method}&Acc&H-score&Acc&H-score&Acc&H-score&Acc&H-score&Acc&H-score \\
					\hline
                    \multirow{6}{*}{\small Single}
                    & FC~\cite{li2019feature} &$63.79$&$55.16$&$54.41$&$52.02$&$41.80$&$41.65$&$44.13$&$43.25$&$51.03\pm{0.24}$&$48.02\pm{0.57}$\\
                    & PAR~\cite{cite:NIPS19PAR} &$65.98$&$57.60$&$55.37$&$54.13$&$41.27$&$41.77$&$42.40$&$42.62$&$51.26\pm{0.27}$&$49.03\pm{0.41}$\\
                    & RSC~\cite{cite:ECCV20RSC} &$60.85$&$53.73$&$54.61$&$54.66$&$38.60$&$38.39$&$44.19$&$44.77$&$49.56\pm{0.44}$&$47.89\pm{0.79}$\\
                    & CuMix~\cite{cite:ECCV20Cumix} &$64.63$&$58.02$&$57.74$&$55.79$&$41.54$&$\cellcolor{yellow!25}43.07$&$42.76$&$40.72$&$51.67\pm{0.12}$&$49.40\pm{0.27}$\\
                    & LP-FT~\cite{kumar2022finetuning} & $\cellcolor{orange!25}68.44$ & $\cellcolor{orange!25}61.74$ & $\cellcolor{yellow!25}60.53$ & $\cellcolor{yellow!25}57.04$ & $\cellcolor{yellow!25}43.56$ & $42.24$ & $\cellcolor{yellow!25}53.07$ & $\cellcolor{yellow!25}50.07$ & $\cellcolor{yellow!25}56.40\pm{0.42}$ & $\cellcolor{yellow!25}52.77\pm{0.76}$ \\
                    & RPF (Ours)& $\cellcolor{red!25}\mathbf{70.67}$ & $\cellcolor{red!25}\mathbf{61.92}$ & $\cellcolor{red!25}\mathbf{63.51}$ & $\cellcolor{red!25}\mathbf{59.55}$ & $\cellcolor{orange!25}44.55$ & $\cellcolor{red!25}43.55$ & $\cellcolor{red!25}\mathbf{56.05}$ & $\cellcolor{red!25}\mathbf{53.02}$ & $\cellcolor{red!25}\mathbf{58.70}\pm{0.46}$ & $\cellcolor{red!25}\mathbf{54.51}\pm{0.57}$ \\
					\hline
                    \multirow{1}{*}{\small Ensemble}
                    & DAML~\cite{shu2021open} & $\cellcolor{yellow!25}65.99$ & $\cellcolor{yellow!25}60.13$ & $\cellcolor{orange!25}61.54$ & $\cellcolor{orange!25}59.00$ & $\cellcolor{red!25}45.13$&$\cellcolor{orange!25}43.12$ & $\cellcolor{orange!25}53.13$ & $\cellcolor{orange!25}51.11$ & $\cellcolor{orange!25}56.45\pm{0.21}$ & $\cellcolor{orange!25}53.34\pm{0.45}$\\
                    \hline
				\end{tabular}%
		}
	\end{center}
        \caption{Results on Office-Home dataset in the open-domain setting.}
        \label{table:office-home}
\end{table*}

\begin{table*}[t]
        \small
	\begin{center}
		\resizebox{1.0\textwidth}{!}{
				\begin{tabular}{ll|cccccccc|cc}
					\hline
					& &\multicolumn{2}{c}{\textbf{Clipart}}&\multicolumn{2}{c}{\textbf{Real}}&\multicolumn{2}{c}{\textbf{Painting}}&\multicolumn{2}{c|}{\textbf{Sketch}}&\multicolumn{2}{c}{\textbf{Avg}} \\
					& \textbf{Method}&Acc&H-score&Acc&H-score&Acc&H-score&Acc&H-score&Acc&H-score \\
					\hline
                    \multirow{6}{*}{\small Single} 
                    & FC~\cite{li2019feature} &$29.91$&$35.42$&$64.77$&$63.65$&$44.13$&$\cellcolor{yellow!25}50.07$&$28.56$&$34.10$&$41.84\pm{0.73}$&$45.81\pm{0.69}$\\
                    & PAR~\cite{cite:NIPS19PAR} &$29.29$&$39.99$&$64.09$&$62.59$&$42.36$&$46.37$&$\cellcolor{yellow!25}30.21$&$\cellcolor{orange!25}39.96$&$41.49\pm{0.63}$&$47.23\pm{0.55}$\\
					& RSC~\cite{cite:ECCV20RSC} &$27.57$&$34.98$&$60.36$&$60.02$&$37.76$&$42.21$&$26.21$&$30.44$&$37.98\pm{0.77}$&$41.91\pm{1.28}$\\
					& CuMix~\cite{cite:ECCV20Cumix} &$30.03$&$\cellcolor{yellow!25}40.18$&$64.61$&$65.07$&$44.37$&$48.70$&$29.72$&$33.70$&$42.18\pm{0.45}$&$\cellcolor{yellow!25}46.91\pm{0.40}$\\
                    & LP-FT~\cite{kumar2022finetuning} & $\cellcolor{yellow!25}33.70$ & $36.74$ & $\cellcolor{red!25}\mathbf{69.84}$ & $\cellcolor{yellow!25}66.34$ & $\cellcolor{yellow!25}47.24$ & $49.29$ & $29.74$ & $33.46$ & $\cellcolor{yellow!25}45.13\pm{0.20}$ & $46.46\pm{0.29}$ \\
                    & RPF (Ours)& $\cellcolor{orange!25}36.28$ & $\cellcolor{orange!25}40.54$ & $\cellcolor{orange!25}69.71$ & $\cellcolor{orange!25}67.08$ & $\cellcolor{red!25}\mathbf{50.69}$ & $\cellcolor{red!25}\textbf{53.01}$ & $\cellcolor{orange!25}31.75$ & $\cellcolor{yellow!25}35.79$ & $\cellcolor{red!25}\mathbf{47.11}\pm{0.61}$ & $\cellcolor{orange!25}49.11\pm{0.56}$ \\
                    \hline
                    \multirow{1}{*}{\small Ensemble} 
                    & DAML~\cite{shu2021open} &$\cellcolor{red!25}\mathbf{37.62}$&$\cellcolor{red!25}\mathbf{44.27}$&$\cellcolor{yellow!25}66.54$&$\cellcolor{red!25}\mathbf{67.80}$&$\cellcolor{orange!25}47.80$&$\cellcolor{orange!25}52.93$&$\cellcolor{red!25}\mathbf{34.48}$&$\cellcolor{red!25}\mathbf{41.82}$&$\cellcolor{orange!25}46.61\pm{0.59}$&$\cellcolor{red!25}\mathbf{51.71}\pm{0.52}$\\
					\hline
				\end{tabular}
		}
	\end{center}
        \caption{Results on Multi-Datasets scenario in the open-domain setting.}
        \label{table:multidataset}
\end{table*}

\subsection{Ablation Study}
\label{subsec:ablation}
We ablate each component of our method to show its effectiveness. Table~\ref{table:abl} shows our ablation study on the Office-Home dataset. HR and FR denotes $\mathcal{L}_{\text{hr}}$ and $\mathcal{L}_{\text{fr}}$ respectively. 
The table clearly indicates that the exclusion of our proposed regularization terms - resulting in training solely with $\mathcal{L}_{\text{lp-ft}}$ (I) - leads to a decrease in both Acc and H-score. Using only one regularization term (II and III) leads to only a marginal gain compared to LP-FT (I) in most cases, and in some target domains, the performance actually degrades. On the other hand, when the two terms are used together (VII - RPF), the performance gain is marked. This implies that the two terms create a synergetic effect for boosting the ODG performance. Next, we compare some possible variants of our method to justify our key design choices. First, without the pre-trained head by linear-probing (IV), our method shows poor performance which indicates that the pre-trained head is inevitable for our method to be effective, bolstering our motivation of not distorting the pre-trained features too much. \emph{HR$_{f}$} (V) maximizes entropy on the features from the target feature extractor $f$, not the frozen pre-trained one $\bar{f_0}$ i.e. $z^i_j$ in (\ref{eq:hr}) is $h^{lp}(f(x^s))$ not $h^{lp}(\bar{f_0}(x^s))$. This leads to improvement in some target domains but its performance is evidently lower than RPF (VII). \emph{HR$_{f}$} does not show much improvement because its objective contradicts with $\mathcal{L}_{\text{lp-ft}}$ which minimizes the cross-entropy loss on features from $f$. A similar trend is also observed in \emph{Ent-Min HR} (VI) which minimizes (as opposed to maximizing) entropy of logit conditioned on the pre-trained features.
\emph{Ent-Min HR} (VI) is also not effective because it decreases the adaptability of the model by fitting the head more to the pre-trained features of the source domains which encourages the head to remain at $h^{lp}$. These two observations justify our method of maximizing entropy of logit conditioned on the pre-trained features.

\subsection{Analysis} \label{analysis}

We analyze the impact of our proposed method on feature-level and logit-level and examine our head's similarity with the pre-trained head using the Office-Home dataset.

\noindent \textbf{Quantitative Analysis on Feature\ } Table~\ref{table:f_anal} shows our  feature level analysis. All the numbers are mean squared error distances. LP-FT denotes model I in Table\ref{table:abl}. Source domain inputs are from the respective validation set. `Domain gap' is the distance between the feature centroids of the same class from different domains (i.e. the centroid for class $c$ of domain $i$ is $\frac{1}{N^i_c} \sum^{N^i_c}_j (f(x^i_j))$). 
The first row is the mean of the domain gap between every possible pair of the source and the target while the second row is only between the sources. Our RPF shows shorter distances than LP-FT which means that the same class centroids from different domains are nearby, indicating a smaller domain gap. `Intra-class dist.' is the average distance between each feature of input and its corresponding centroid. It is measured for each class and then averaged over the classes. This metric signifies the larger the value, the less the features are clustered by class or highly dispersed. For both the target and the source domains, ours is smaller, indicating that our method reduces the intra-class variance. `Diff. from $f_0$' shows how each feature of input is changed from its pre-trained feature after training, hence showing $||f(x) - f_0(x)||^2_2$. It shows the average of  difference over entire inputs.
We observe that regardless of the domain, features of RPF are changed less from their initial pre-trained features compared to LP-FT, yet RPF shows higher performance. We conjecture that $f$ is changed less from $f_0$ due to our $\mathcal{L}_{\text{fr}}$ which helps to preserve the pre-trained features, not distort them excessively but rather utilize them to generalize better to unseen domains. Overall, with our proposed method, the domain gap is narrowed, intra-class variance is reduced and the pre-trained features are more preserved in favor of better generalization to unseen domains.

\noindent \textbf{Quantitative Analysis on Logit \ } Table~\ref{table:l_anal} analyzes the maximum confidence score ($\max\limits_{c\in C} \sigma(z^t)^c$) and the entropy of the logits given the target domain inputs. Since the input is the target domain, open (unknown) classes are included, so we distinguish the two types in the table. The average over the entire target samples is reported. As shown in the table, RPF produces logits with a lower maximum confidence score and higher entropy for both known and unknown classes over the four target domains, meaning that the class probability distribution of $\sigma(z)$ is smoother. It is noteworthy that even though RPF shows higher entropy, it achieves better performance than LP-FT. RPF trains the model not to predict too over-confidently on the unknown classes and prevents it from being biased towards the source samples during training. We conjecture that this is possible because $\mathcal{L}_{\text{fr}}$ regulates $f$ to stay similar to $f_0$ while $\mathcal{L}_{\text{hr}}$ penalizes the head to deviate from $h^{lp}$ to output high entropy on the pre-trained features of the source domain inputs.

\begin{figure}[t]
    \centering
    \includegraphics[width = 0.95\linewidth]{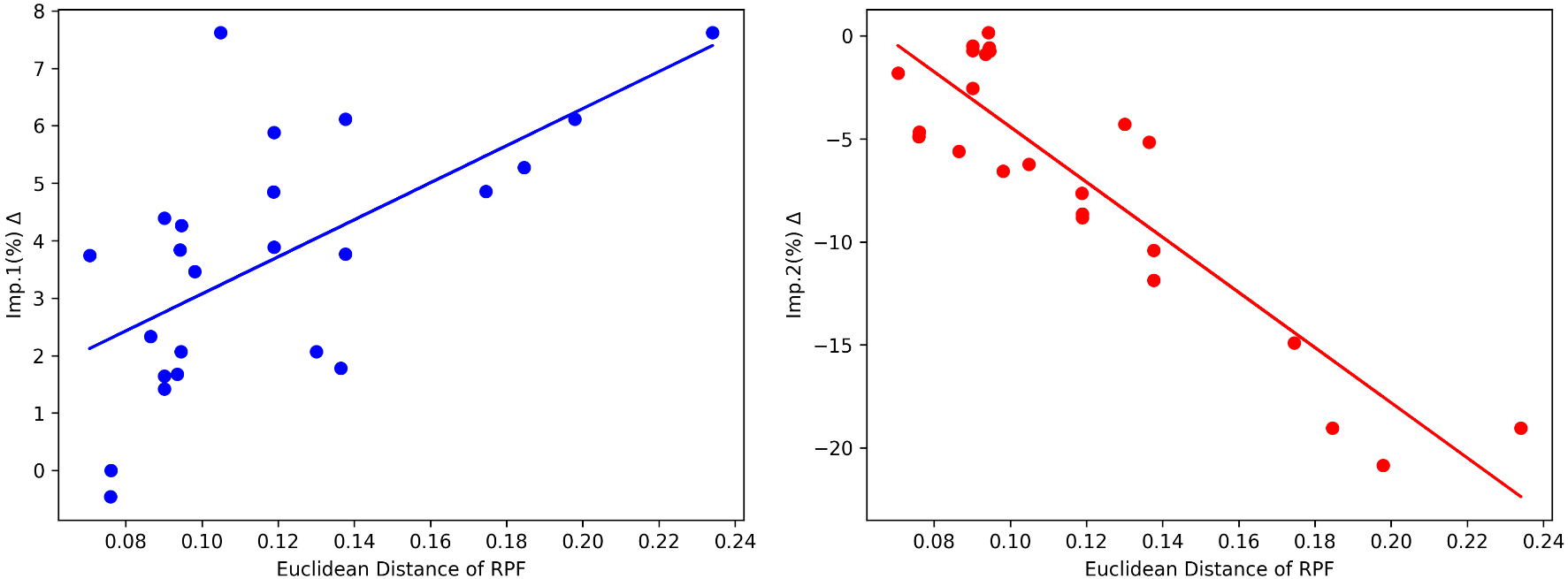}
    \caption{ The correlation between the Euclidean distance of RPF and the $\Delta$ of Imp.1 (Left) and Imp.2 (Right) respectively.}
    \label{fig:correlation}
\end{figure}

\noindent \textbf{Head parameters \ } In Table~\ref{table:lp_compare}, we analyze  and observe a meaningful correlation between the distance of the head parameters from that of the linear-probing model and the performance.
As explained earlier, our method requires linear-probing the head in advance with a fixed pre-trained feature extractor $\bar{f_0}(.)$. We analyze how our method affects the head in the parameter space by measuring its Euclidean distance to the pre-trained head, $h^{lp}$.
Since the head is a $C$-way classifier, the shape of its weights and bias are $\mathbb{R}^{C \times d}$ and $\mathbb{R}^C$ respectively, where $d$ is the dimension of the feature, $f(x) \in \mathbb{R}^d$. We concatenate the weights and the bias term, creating $W \in \mathbb{R}^{C \times d+1}$, then measure the Euclidean distance between $W$ of the head trained by each method and that of the $h^{lp}$. The Euclidean distance is calculated row-wise (class-wise) for $W$ and then averaged over the class dimension. In the table, the head of RPF shows a larger Euclidean distance from $h^{lp}$ than LP-FT which means that RPF has changed the head parameters more from its initial weight $h^{lp}$. 
Imp.1(\%) is the performance improvement ratio of Acc compared to the LP model, $((\frac{Model Acc}{LP Acc}-1)\times100)$. RPF consistently achieves higher improvement from the LP model compared to LP-FT.
Imp.2(\%) shows how much improvement is achieved only by changing the head with the one trained by each method i.e. improvement of $h^{*}(\bar{f_0}(.))$ compared to $h^{lp}(\bar{f_0}(.))$ where $h^{*}$ is the trained head by each method. 
RPF performs worse than LP-FT in Imp.2. This is because RPF changes the head more from its initial weight, $h^{lp}$ via $\mathcal{L}_{\text{hr}}$ which maximizes entropy on features from $\bar{f_0}(.)$ while LP-FT does not change $h^{lp}$ much but rather adopts it. \\
\indent $\Delta$ refers to the absolute difference between RPF and LP-FT in the two Imp.'s. To better capture the correlation between $\Delta$ and the Euclidean distance of RPF, we additionally run three more experiments on each domain, hence 24 runs total. The Spearman rank correlation coefficient~\shortcite{ref1} between the Euclidean distance of RPF and $\Delta$ of Imp.1 and Imp.2 are $0.6562$ and $-0.7561$ respectively. Fig.~\ref{fig:correlation} shows the scatter plots of the two correlations for all 24 runs. 
The negative correlation of Imp.2 is expected since the head changes more from $h^{lp}$, it becomes less fitted to correctly classify features from $\bar{f_0}$.
The positive correlation of Imp.1 implies that adjusting the head to deviate from $h^{lp}$ plays a key role in improving the adaptability of the model for unseen domains. We conjecture this is because $\mathcal{L}_{\text{fr}}$ preserves the pre-trained features by clustering them around the $P_c$s and making the head flexible by minimizing $\mathcal{L}_{\text{hr}}$ while minimizing $\mathcal{L}_{\text{lp-ft}}$ as well to ensure the model correctly classifies training samples without compromising its representation. Consequently, the softmax output becomes smoother (Tab.\ref{table:l_anal}) and $\mathcal{L}_{\text{lp-ft}}$ increases (Fig.\ref{fig:train_loss}) which prevents the model from being biased towards the source domains and enhances its adaptability to unseen domains.

\subsection{Comparison with other methods}
In Table~\ref{table:pacs}, \ref{table:office-home}, and \ref{table:multidataset} we compare our method with other methods under the ODG setting. We show the performance on four different target domains for each benchmark and report the average over the four target domains as well. The compared methods are \cite{li2019feature, cite:ECCV20Cumix, cite:NIPS19PAR, cite:ECCV20RSC} taken from \cite{shu2021open}. They are heterogeneous~\cite{li2019feature}, augmentation-based~\cite{cite:ECCV20Cumix} DG methods, and methods that learn robust and generalizable features~\cite{cite:NIPS19PAR, cite:ECCV20RSC}. LP-FT~\cite{kumar2022finetuning} is the direct baseline of our method which shows the results without our proposed regularization terms. The performance gap between LP-FT and RPF presents the effectiveness of our proposed terms. In all the three benchmarks, our method leads to better Acc and H-score. DAML~\cite{shu2021open} is another baseline that initially proposed this ODG setting as explained earlier. Note that DAML shows the performance of an \emph{ensemble} of multiple source domain networks which means that in this case, it is the ensemble of 3 networks, since 3 source domains are used in each benchmark. 
RPF reaches the highest performance on all three benchmarks among the non-ensemble methods when averaged across the target domains. In Office-Home, it even outperforms DAML and also shows the best performance for the H-score of PACS and Acc of Multi-Datasets. These results are quite remarkable considering that the performance of DAML is from the ensembled network.

\section{Conclusion}
We investigate the open domain generalization problem and introduce a method to boost the performance with a single network. Our method adopts a sequential learning scheme of training only the head first and then fine-tuning the whole model. We propose two regularization terms taking advantage of the pre-trained features during fine-tuning for the feature extractor and the head, respectively. We observe our method prevents distorting the pre-trained features excessively while adjusting the head to deviate from the pre-trained head. With extensive experiments, we analyze how our proposed method affects the model and validate its efficacy in improving the adaptability of the model to unseen domains. Our method shows competitive performance against previous leading method, DAML, in several benchmarks, while greatly reducing the inference cost with a single network.

\appendix

\section{Datasets and Class splits}
In this section, we further explain the details of the experimental setting of open domain generalization. Please note that this experimental setting is originally proposed by \cite{shu2021open} and we exactly follow its details for a fair comparison. We clearly note that the following tables~\ref{table:PACSSupp},~\ref{table:OfficeHomeSupp} and ~\ref{table:Multi} are borrowed from the supplementary materials of \cite{shu2021open} and all the credits for proposing the ODG setting and providing details of class splits goes to \cite{shu2021open}.

\textbf{PACS}~\cite{cite:ICCV17DBA} comprises of four distinct image styles, namely photo (\textbf{P}), art-painting (\textbf{A}), cartoon (\textbf{C}), and sketch (\textbf{S}). All four domains share a common label set of 7 classes. Four cross-domain tasks are formed by utilizing each domain as the target domain and the remaining three domains as the source domains: CPS-A, PAC-S, ACS-P, and SPA-C. In order to implement open-domain generalization environment, the label space of the dataset is split, resulting in varying label spaces across different domains. Table \ref{table:PACSSupp} displays the specific categories contained within each domain. Note that the order of the source domains matters. For instance, CPS-A and CSP-A would be considered different because Source-2 and Source-3 have different label sets. The target domain includes all the classes present in the dataset, with class 6 (person) being the open class. 

\begin{table}[b]
	\begin{center}
            \begin{tabular}{cc}
                \hline
                \textbf{Domain}&\textbf{Classes} \\
                \hline
                Source-1 &$3, 0, 1$\\
                Source-2 &$4, 0, 2$\\
                Source-3 &$5, 1, 2$\\
                Target &$0, 1, 2, 3, 4, 5, 6$\\
                \hline
            \end{tabular}
	\end{center}
        \caption{Class split of PACS dataset.} 
        \label{table:PACSSupp}
\end{table}

\textbf{Office-Home}~\cite{cite:CVPR17OfficeHome} dataset is comprised of images from four distinct domains: Real-world (\textbf{R}), Product (\textbf{P}), Clip art (\textbf{C}) and Artistic (\textbf{A}) and consists of 65 classes which makes it more challenging than \textbf{PACS}~\cite{cite:ICCV17DBA}. Similar to \textbf{PACS}, four different open domain generalization tasks: ACP-R, ACR-P, APR-C and CPR-A are formed where each domain is used as the target domain, and the remaining three domains serve as the source domains respectively. The 65 classes are distributed among the four domains to realize the open-domain setting. The order of source domains matters as in \textbf{PACS}, for example, APC-R would be different from ACP-R. Different from \textbf{PACS}, this setting assigns more open classes (54-64) in the target domain and some classes in the source domains are not included in the target domain which makes this ODG setting more intricate.

\begin{table}[t]
	\begin{center}
            \begin{tabular}{cc}
                \hline
                \textbf{Domain}&\textbf{Classes} \\
                \hline
                Source-1 &$0-2, 3-8, 9-14, 21-31$\\
                Source-2 &$0-2, 3-8, 15-20, 32-42$\\
                Source-3 &$0-2, 9-14, 15-20, 43-53$\\
                Target &$0, 3-4, 9-10, 15-16,$ \\
                &$21-23, 32-34, 43-45, 54-64$\\
                \hline
            \end{tabular}
	\end{center}
        \caption{Class split of Office-Home dataset.}
        \label{table:OfficeHomeSupp}
\end{table}

As explained in the main paper, \cite{shu2021open} proposed the \textbf{Multi-Datasets} scenario to realize a more realistic open domain generalization setting in which the source domains are obtained from different resources and the model is trained to show high performance on an unseen target domain. Three public datasets, \textbf{Office-31}~\cite{cite:ECCV10Office}, \textbf{STL-10}~\cite{coates2011analysis} and \textbf{Visda2017}~\cite{visda2017} are used as the source domains and the trained model is evaluated on the four domains (Clipart, Real, Painting and Sketch) of \textbf{DomainNet}~\cite{peng2019moment}. \textbf{Office-31} contains 31 classes in three domains: Amazon, DSLR and Webcam. In this setting, the Amazon domain is used. It consists of objects commonly encountered in an office environment. \textbf{STL-10} consists of 10 common objects and its labeled data is used as one of the source domains. \textbf{Visda2017} is a simulation-to-real dataset for domain adaptation with 12 categories. Its train set is utilized as one of the source domains which is synthetic 2D images of 3D models generated from different angles and lighting conditions. \textbf{DomainNet} is another dataset for a cross-domain generalization benchmark. It has 345 classes shared across the four domains. Since there exist too many classes in each domain, 23 classes that exist in the union of the three source domain classes are preserved as the known classes, and other 20 classes are sampled as the unknown classes. Similar to \textbf{Office-Home}, some classes in the source domains are not included in the target domain. Since there is a large domain and label set discrepancy between the sources and the target, this scenario naturally forms a decent open-domain generalization scenario.

\begin{table}[t]
	\begin{center}
            \begin{tabular}{cc}
                \hline
                \textbf{Domain}&\textbf{Classes} \\
                \hline
                Office-31 &$0-30$\\
                Visda &$1, 31-41$\\
                STL-10 &$31, 33, 34, 41, 42-47$\\
                DomainNet &$0, 1, 5, 6, 10, 11, 14, 17, 20, 26$ \\
                &$31-36, 39-43, 45-46, 48-67$\\
                \hline
            \end{tabular}
	\end{center}
        \caption{Class split of Multi-Datasets scenario.}
        \label{table:Multi}
\end{table}

\begin{table*}[t]
        \small
	\begin{center}
		\resizebox{0.9\textwidth}{!}{
				\begin{tabular}{lcccccccccc}
					\hline
					&\multicolumn{2}{c}{\textbf{Real-World}}&\multicolumn{2}{c}{\textbf{Product}}&\multicolumn{2}{c}{\textbf{Clipart}}&\multicolumn{2}{c}{\textbf{Art}}&\multicolumn{2}{c}{\textbf{Avg}} \\
					\textbf{$\lambda_{hr}$}&Acc&H-score&Acc&H-score&Acc&H-score&Acc&H-score&Acc&H-score \\
					\hline
                    \emph{1.0} & $64.66$ & $58.91$ & $57.90$ & $54.87$ & $41.21$ & $41.31$ & $50.45$ & $47.04$ & $53.56\pm{1.26}$ & $50.53\pm{0.49}$ \\
                    
                    \emph{0.5} & $65.73$ & $59.49$ & $60.81$ & $57.77$ & $42.86$ & $43.41$ & $55.40$ & $51.76$ & $56.20\pm{0.60}$ & $53.11\pm{0.49}$ \\
                    
                    0.1 & $\mathbf{70.67}$ & $\mathbf{61.92}$ & $\mathbf{63.51}$ & $\mathbf{59.55}$ & $\mathbf{44.55}$ & $\mathbf{43.55}$ & $\mathbf{56.05}$ & $\mathbf{53.02}$ & $\mathbf{58.70}\pm{0.46}$ & $\mathbf{54.51}\pm{0.57}$ \\
					\hline
				\end{tabular}%
		}
	\end{center}
	\caption{Ablation study on $\lambda_{hr}$ using the Office-Home dataset under the open-domain setting. All experiments are trained with $\mathcal{L}_{\text{fr}}$.}
        \label{table:abl_lambda}
\end{table*}

\begin{table*}[t]
        \small
	\begin{center}
		\resizebox{0.9\linewidth}{!}{
				\begin{tabular}{lc|ccc|ccc|ccc|ccc}
					\hline
					\textbf{Metric} & \textbf{Target domain} & \multicolumn{3}{c|}{\textbf{Real-World}} & \multicolumn{3}{c|}{\textbf{Product}} & \multicolumn{3}{c|}{\textbf{Clipart}} & \multicolumn{3}{c}{\textbf{Art}}  \\
                     & \textbf{Source domains} & \textbf{A} & \textbf{C} & \textbf{P} & \textbf{A} & \textbf{C} & \textbf{R} & \textbf{A} & \textbf{P} & \textbf{R} & \textbf{C} & \textbf{P} & \textbf{R} \\
					\hline
					\multirow{2}{*}{Domain gap} 
                            & RPF & $0.17$ & $0.18$ & $0.13$ & $0.17$ & $0.18$ & $0.12$ & $0.24$ & $0.30$ & $0.28$ & $0.11$ & $0.18$ & $0.16$ \\
                            & LP-FT & $0.19$ & $0.23$ & $0.16$ & $0.26$ & $0.30$ & $0.22$ & $0.26$ & $0.33$ & $0.30$ & $0.20$ & $0.26$ & $0.21$ \\ 
                    \hline
                    \multirow{2}{*}{Intra-class dist.} 
                            & RPF & $0.32$ & $0.37$ & $0.36$ & $0.23$ & $0.26$ & $0.28$ & $0.36$ & $0.40$ & $0.43$ & $0.28$ & $0.28$ & $0.30$ \\
                            & LP-FT & $0.39$ & $0.46$ & $0.43$ & $0.39$ & $0.46$ & $0.45$ & $0.40$ & $0.44$ & $0.47$ & $0.46$ & $0.42$ & $0.45$ \\
                    \hline
				\end{tabular}%
		}
	\end{center}
	\caption{Analysis on domain gap and intra-class distance for each source-target pair and each source domain.}
        \label{table:f_anal_supple}
\end{table*}

Once again, we notify that all the contributions and credits for proposing the above ODG setting and the detailed class splits should be given to \cite{shu2021open}. For the train and validation split of the source domains, \cite{shu2021open} explains that it follows the protocol in \cite{cite:ICCV17DBA} for \textbf{PACS} and for other datasets, it randomly selects $10\%$ of data in each category of the source domains as their validation sets. 

\section{Implementation Details}

Experiments are conducted using a ResNet-18~\cite{he2016deep} backbone pre-trained on ImageNet~\cite{deng2009imagenet} with a head whose final output dimension is defined as the number of classes in the source domains. We train the model for 30 epochs using the SGD optimizer with an initial learning rate of 0.001 and batch size of 32. The learning rate is decreased after 24 epochs by a factor of 10. We choose the model that shows the highest accuracy on the held-out source validation set and evaluate it on the target domain. We use the Pytorch~\cite{NEURIPS2019_9015} automatic differentiation framework and a NVIDIA RTX 3090 for conducting experiments. We report the average of 3 runs for each experiment. Two metrics are used for the evaluation: 1) Acc, which is the accuracy only on the known classes (open classes are not given when calculating Acc) and 2) H-score, which is the harmonic mean of known and open class accuracy (both known and open classes are given) following ~\cite{Fu_2020_ECCV, shu2021open}.

\section{Ablation study on $\lambda_{hr}$}

In the main paper, we mention that a large $\lambda_{hr}$ leads to the impairment of the head, thus deteriorating the performance. Table.~\ref{table:abl_lambda} shows our ablation study on $\lambda_{hr}$. As $\lambda_{hr}$ increases, both Acc and H-score decrease in all four target domains. Since $\mathcal{L}_{\text{fr}}$ regulates the $f$ to stay similar to $f_0$ while $\mathcal{L}_{\text{hr}}$ regulates $h$ to deviate from $h^{lp}$, if $\mathcal{L}_{\text{hr}}$ is too strong, the head will deviate too far from $h^{lp}$ and contradicts with $\mathcal{L}_{\text{lp-ft}}$ as well which results in a sub-optimal solution that can not correctly classify the features from the $f$. 

\section{More results on domain gap and intra-class distance analysis}

In the main paper, we show the average domain gap between every possible pair of the source and the target for each ODG task in the Office-Home dataset. We also report the intra-class distance averaged over the source domains. Therefore, in Table.~\ref{table:f_anal_supple}, we show the domain gap for each source-target pair and intra-class distance for each source domain. In the table, we divide the ODG task via vertical lines. `Domain gap' is the mean squared distance between feature centroids of the target and the source domain while `Intra-class dist.' refers to the mean distance between each feature point to its centroid feature for each source domain. As we already observed in the main paper, PRF shows smaller domain gap and intra-class distance which indicates that our proposed regularization terms indeed help to narrow the domain gap and cluster the features by classes.

\begin{figure*}[t]
    \centering
    \includegraphics[width = 0.9 \linewidth]{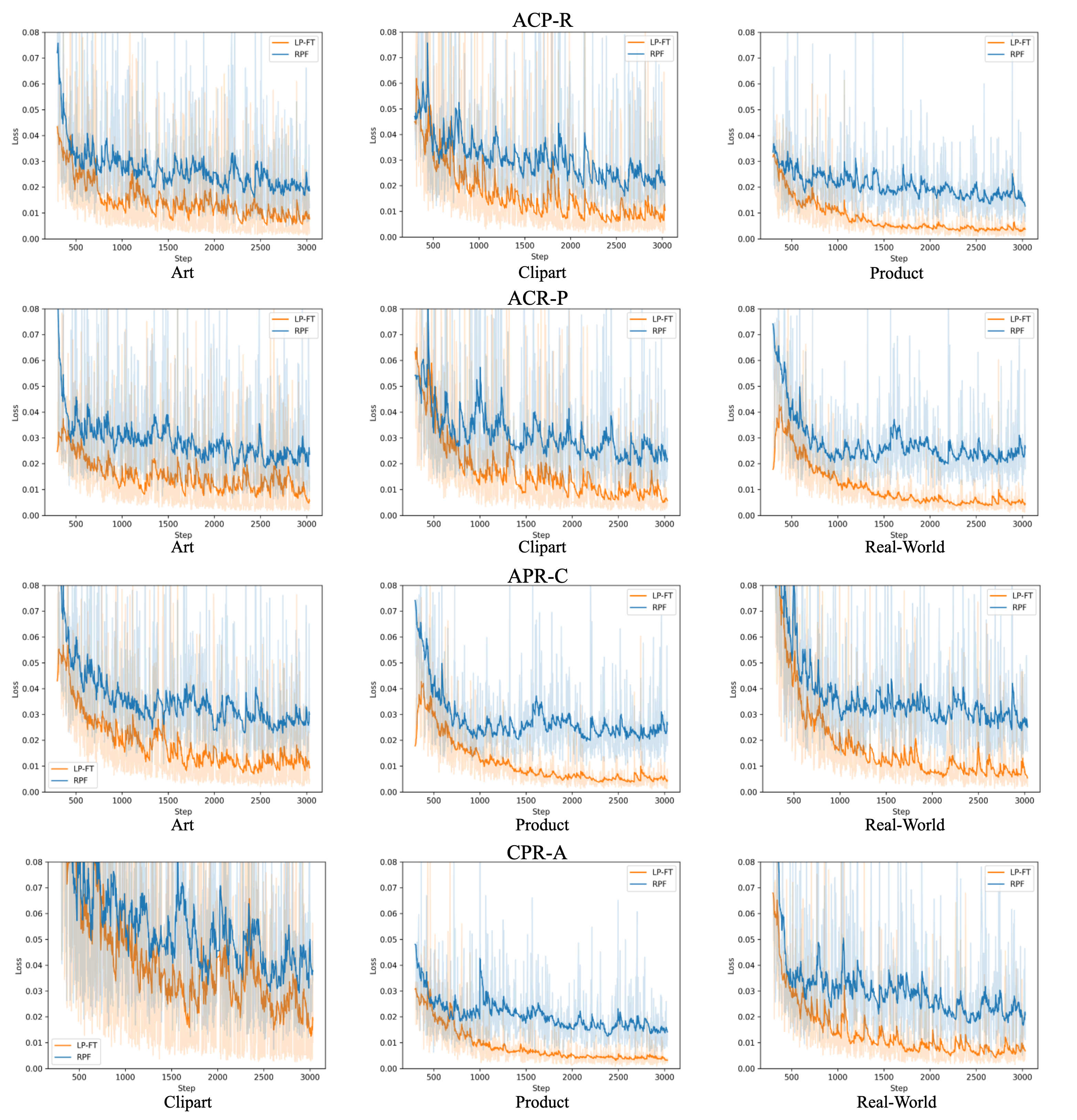}
    \caption{The scale of $\mathcal{L}_{\text{lp-ft}}$ for different ODG task in the Office-Home dataset.}
    \label{fig:loss_plot_supple}
\end{figure*}

\section{$\mathcal{L}_{\text{lp-ft}}$ comparison between LP-FT and RPF}

Fig.~\ref{fig:loss_plot_supple} shows the scale of $\mathcal{L}_{\text{lp-ft}}$ on each source domain in each ODG task. Each row of the figure refers to a different ODG task: ACP-R, ACR-P, APR-C and CPR-A. Plots with higher opacity are smoothed plots using the exponential moving average. As shown in the table, RPF shows a higher loss scale than LP-FT on each source domain of each ODG task. Even though the two methods show different loss scales, they both converge to 100\% train accuracy. RPF showing a higher cross-entropy loss on the source domain samples seen during training implies that the model is less overfitted to the source domains and its predicted class probability distribution is more smooth rather than peaky compared to LP-FT. We believe that RPF learns a more generalizable representation because it is less biased toward the source domains.

\begin{figure*}[t]
    \centering
    \includegraphics[width = 1.0 \linewidth]{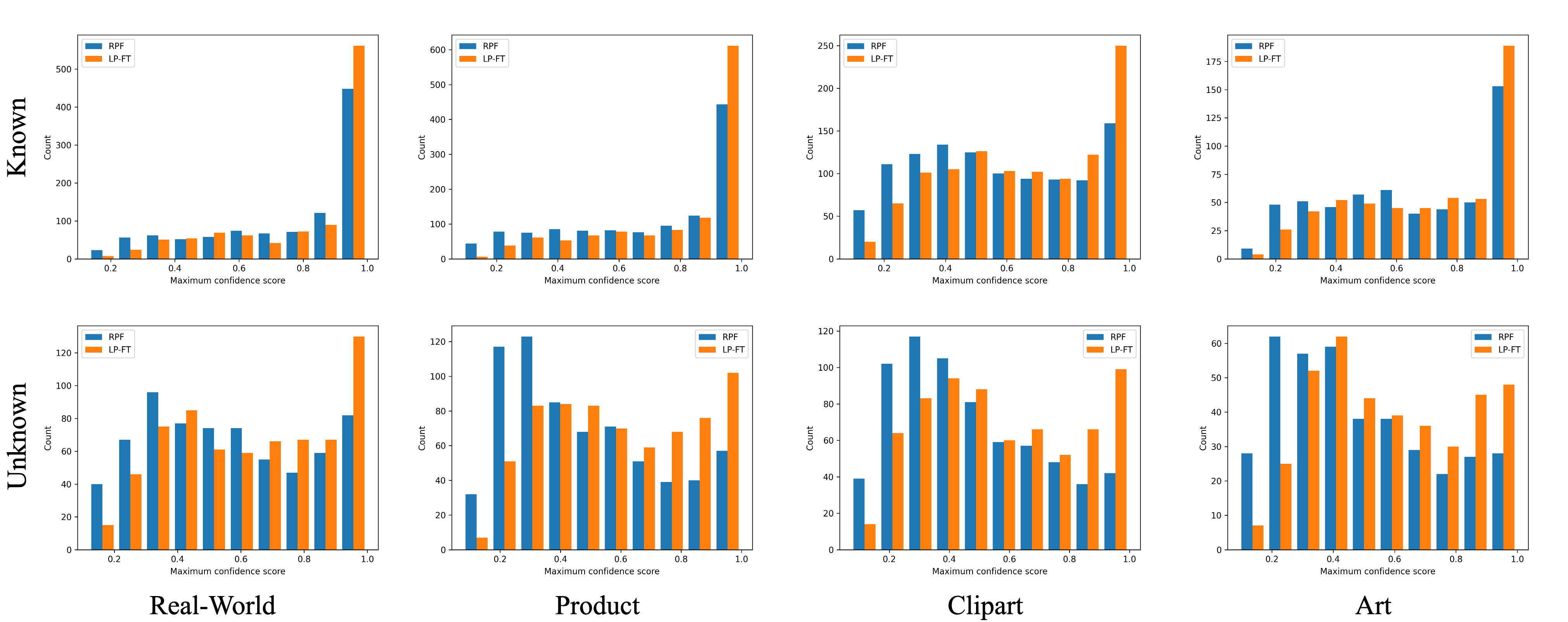}
    \caption{Histograms of known and unknown classes on the four target domains of the Office-Home dataset.}
    \label{fig:histogram_officehome_supple}
\end{figure*}

\begin{figure*}[t]
    \centering
    \includegraphics[width = 1.0 \linewidth]{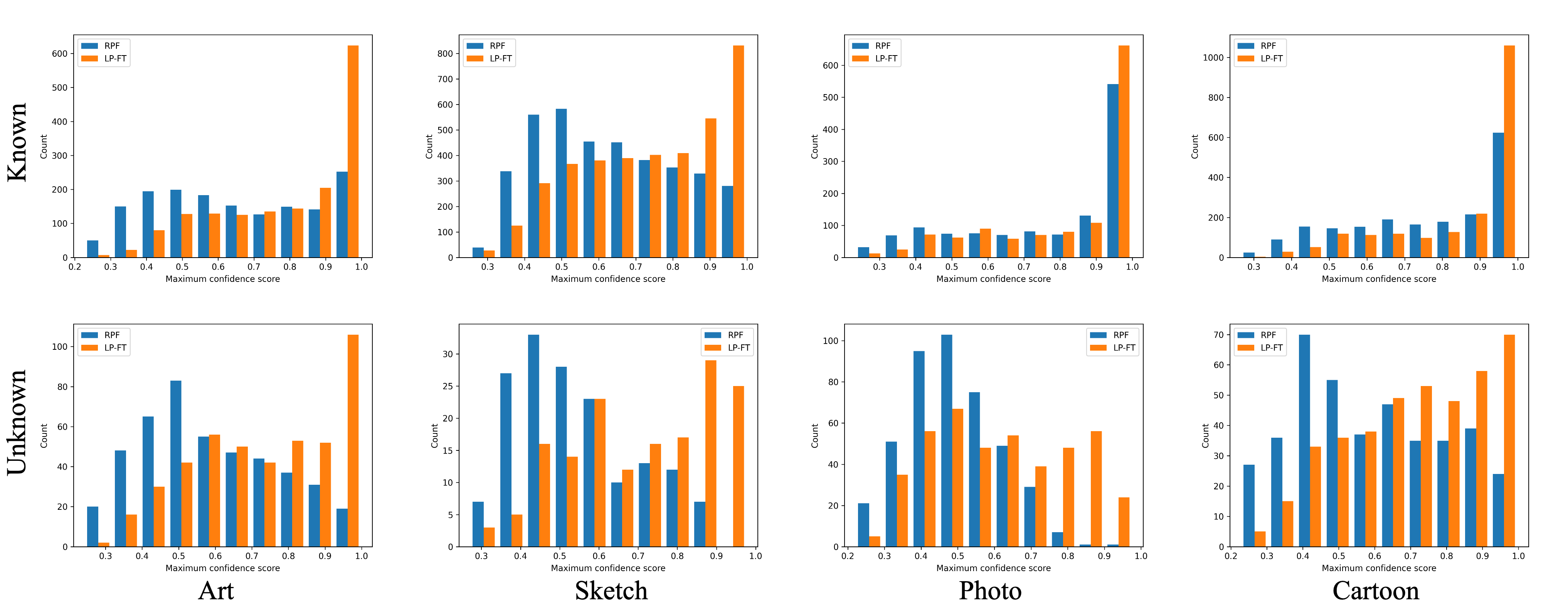}
    \caption{Histograms of known and unknown classes on the four target domains of the PACS dataset.}
    \label{fig:histogram_pacs_supple}
\end{figure*}

\begin{figure*}[t]
    \centering
    \includegraphics[width = 1.0 \linewidth]{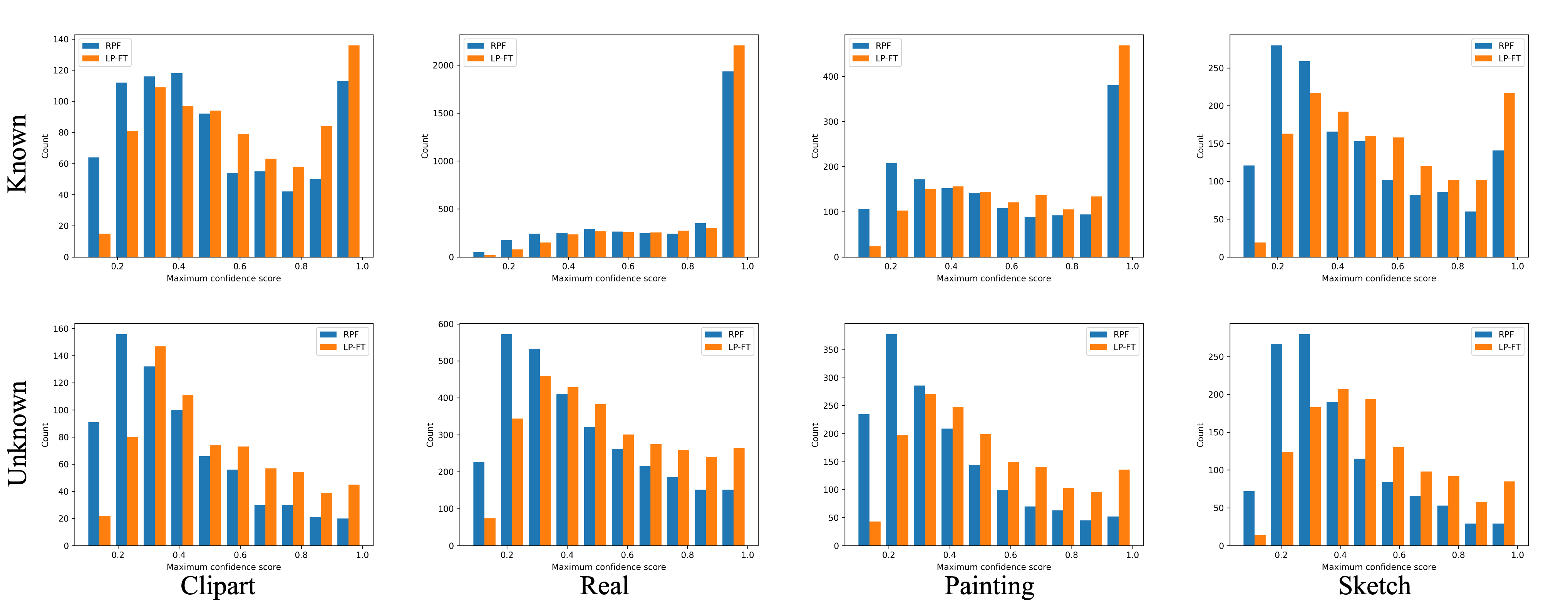}
    \caption{Histograms of known and unknown classes on the four target domains of the Multi-Datasets scenario.}
    \label{fig:histogram_multi_supple}
\end{figure*}

\begin{figure*}[t]
    \centering
    \includegraphics[width = 1.0 \linewidth]{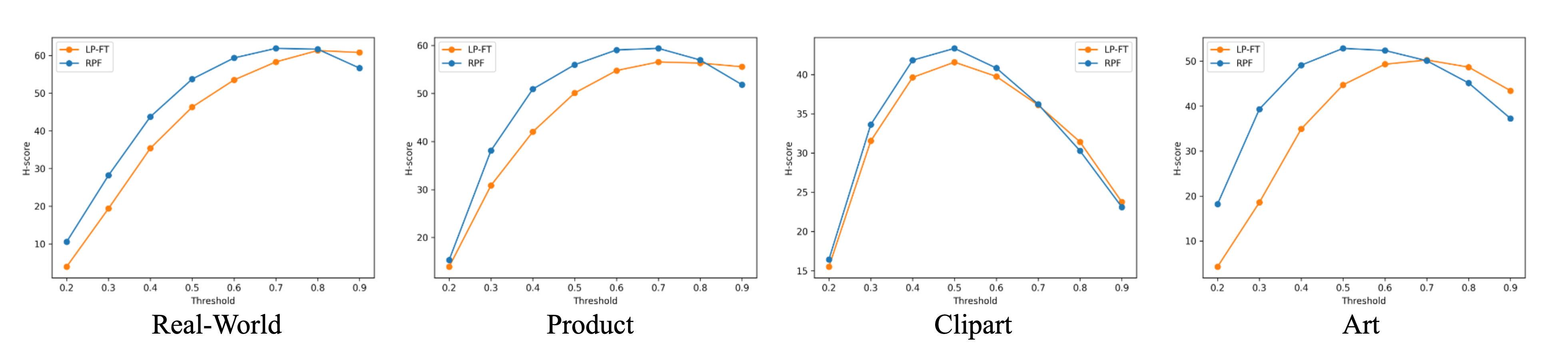}
    \caption{Threshold robustness analysis on the the Office-Home dataset.}
    \label{fig:office_home_threshold_supple}
\end{figure*}

\begin{figure*}[t]
    \centering
    \includegraphics[width = 1.0 \linewidth]{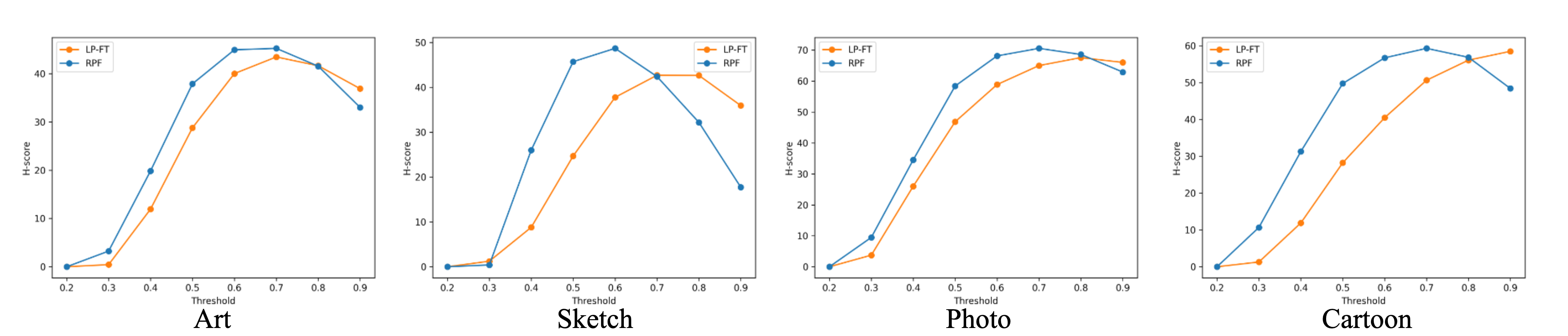}
    \caption{Threshold robustness analysis on the the PACS dataset.}
    \label{fig:pacs_threshold_supple}
\end{figure*}

\begin{figure*}[t]
    \centering
    \includegraphics[width = 1.0 \linewidth]{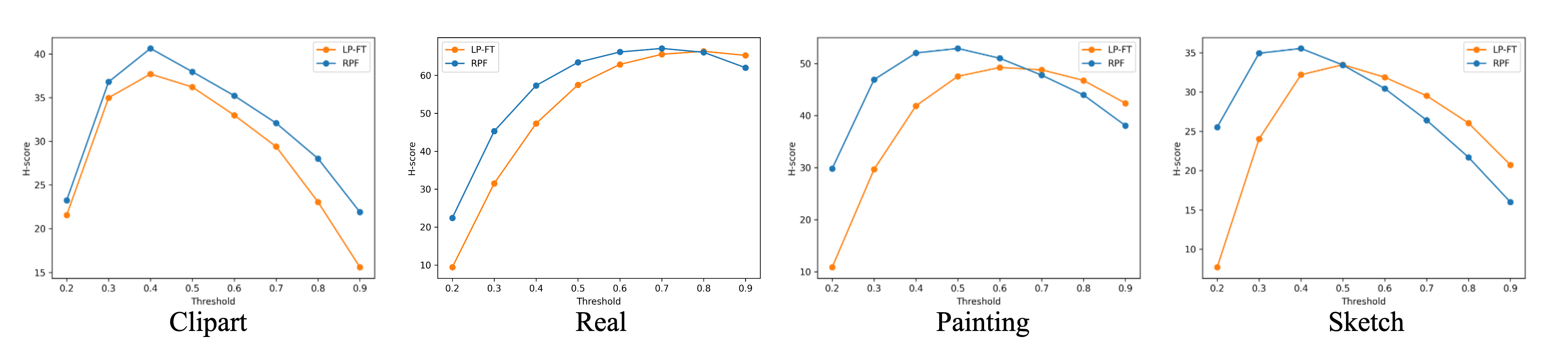}
    \caption{Threshold robustness analysis on the the Multi-Datasets scenario.}
    \label{fig:multi_threshold_supple}
\end{figure*}

\section{Maximum confidence score histograms} 

Fig.~\ref{fig:histogram_officehome_supple}~\ref{fig:histogram_pacs_supple} and ~\ref{fig:histogram_multi_supple} show the histograms of maximum confidence score for known and unknown classes over the four target domains of each ODG benchmark. A similar trend is observed from all three benchmarks, which is that RPF outputs a lower maximum confidence score on the unknown classes compared to LP-FT. RPF also shows a relatively lower maximum confidence score on known classes as well. LP-FT exhibits a tendency of showing overly confident predictions for both known and unknown classes while RPF shows rather under-confident predictions. It indicates that the predicted softmax output of RPF has more smooth distribution and higher entropy. Higher entropy implies that the model is not too biased towards the source domains samples and classes seen during training.

\section{Threshold robustness analysis}

For a fair comparison, we follow the evaluation protocol of \cite{shu2021open} which is to measure the H-score using the threshold that shows the best score. Therefore, in Fig.~\ref{fig:pacs_threshold_supple}~\ref{fig:office_home_threshold_supple} and ~\ref{fig:multi_threshold_supple}, we present the H-score of both RPF and LP-FT using 8 different thresholds with equal interval to show the robustness to change in the threshold of each method. The results are the average of three experiments. We observe that RPF achieves a better H-score than LP-FT on most of the thresholds while it shows lower H-scores on high thresholds. The reason why RPF shows lower H-scores on high thresholds is because it predicts less confidently on the known classes as shown in the previous section, thus classifying most of the known classes as unknowns with high thresholds.

\bibliography{aaai24}

\begin{thebibliography}{44}
\providecommand{\natexlab}[1]{#1}

\bibitem[{ref(2008)}]{ref1}
 2008.
\newblock \emph{Spearman Rank Correlation Coefficient}, 502--505.
\newblock New York, NY: Springer New York.
\newblock ISBN 978-0-387-32833-1.

\bibitem[{Cha et~al.(2021)Cha, Chun, Lee, Cho, Park, Lee, and Park}]{cha2021swad}
Cha, J.; Chun, S.; Lee, K.; Cho, H.-C.; Park, S.; Lee, Y.; and Park, S. 2021.
\newblock Swad: Domain generalization by seeking flat minima.
\newblock \emph{Advances in Neural Information Processing Systems}, 34: 22405--22418.

\bibitem[{Cha et~al.(2022)Cha, Lee, Park, and Chun}]{cha2022domain}
Cha, J.; Lee, K.; Park, S.; and Chun, S. 2022.
\newblock Domain Generalization by Mutual-Information Regularization with Pre-trained Models.
\newblock \emph{arXiv preprint arXiv:2203.10789}.

\bibitem[{Coates, Ng, and Lee(2011)}]{coates2011analysis}
Coates, A.; Ng, A.; and Lee, H. 2011.
\newblock An analysis of single-layer networks in unsupervised feature learning.
\newblock In \emph{International Conference on Artificial Intelligence and Statistics (AISTATS)}, 215--223.

\bibitem[{Deng et~al.(2009)Deng, Dong, Socher, Li, Li, and Fei-Fei}]{deng2009imagenet}
Deng, J.; Dong, W.; Socher, R.; Li, L.-J.; Li, K.; and Fei-Fei, L. 2009.
\newblock Imagenet: A large-scale hierarchical image database.
\newblock In \emph{2009 IEEE conference on computer vision and pattern recognition}, 248--255. Ieee.

\bibitem[{Dou et~al.(2019)Dou, Coelho~de Castro, Kamnitsas, and Glocker}]{dou2019domain}
Dou, Q.; Coelho~de Castro, D.; Kamnitsas, K.; and Glocker, B. 2019.
\newblock Domain generalization via model-agnostic learning of semantic features.
\newblock \emph{Advances in Neural Information Processing Systems}, 32.

\bibitem[{Fu et~al.(2020)Fu, Cao, Long, and Wang}]{Fu_2020_ECCV}
Fu, B.; Cao, Z.; Long, M.; and Wang, J. 2020.
\newblock Learning to Detect Open Classes for Universal Domain Adaptation.
\newblock In \emph{European Conference on Computer Vision (ECCV)}.

\bibitem[{Ghifary et~al.(2016)Ghifary, Balduzzi, Kleijn, and Zhang}]{ghifary2016scatter}
Ghifary, M.; Balduzzi, D.; Kleijn, W.~B.; and Zhang, M. 2016.
\newblock Scatter component analysis: A unified framework for domain adaptation and domain generalization.
\newblock \emph{IEEE transactions on pattern analysis and machine intelligence}, 39(7): 1414--1430.

\bibitem[{Ghifary et~al.(2015)Ghifary, Kleijn, Zhang, and Balduzzi}]{ghifary2015domain}
Ghifary, M.; Kleijn, W.~B.; Zhang, M.; and Balduzzi, D. 2015.
\newblock Domain generalization for object recognition with multi-task autoencoders.
\newblock In \emph{Proceedings of the IEEE international conference on computer vision}, 2551--2559.

\bibitem[{He et~al.(2016)He, Zhang, Ren, and Sun}]{he2016deep}
He, K.; Zhang, X.; Ren, S.; and Sun, J. 2016.
\newblock Deep residual learning for image recognition.
\newblock In \emph{IEEE Conference on Computer Vision and Pattern Recognition (CVPR)}, 770--778.

\bibitem[{Howard and Ruder(2018)}]{howard2018universal}
Howard, J.; and Ruder, S. 2018.
\newblock Universal Language Model Fine-tuning for Text Classification.
\newblock In \emph{Proceedings of the 56th Annual Meeting of the Association for Computational Linguistics (Volume 1: Long Papers)}, 328--339.

\bibitem[{Huang et~al.(2020)Huang, Wang, Xing, and Huang}]{cite:ECCV20RSC}
Huang, Z.; Wang, H.; Xing, E.~P.; and Huang, D. 2020.
\newblock Self-Challenging Improves Cross-Domain Generalization.
\newblock In \emph{European Conference on Computer Vision (ECCV)}.

\bibitem[{Izmailov et~al.(2018)Izmailov, Podoprikhin, Garipov, Vetrov, and Wilson}]{izmailov2018averaging}
Izmailov, P.; Podoprikhin, D.; Garipov, T.; Vetrov, D.; and Wilson, A.~G. 2018.
\newblock Averaging weights leads to wider optima and better generalization.
\newblock \emph{arXiv preprint arXiv:1803.05407}.

\bibitem[{Kanavati and Tsuneki(2021)}]{kanavati2021partial}
Kanavati, F.; and Tsuneki, M. 2021.
\newblock Partial transfusion: on the expressive influence of trainable batch norm parameters for transfer learning.
\newblock In \emph{Medical Imaging with Deep Learning}, 338--353. PMLR.

\bibitem[{Kim et~al.(2021)Kim, Yoo, Park, Kim, and Lee}]{kim2021selfreg}
Kim, D.; Yoo, Y.; Park, S.; Kim, J.; and Lee, J. 2021.
\newblock Selfreg: Self-supervised contrastive regularization for domain generalization.
\newblock In \emph{Proceedings of the IEEE/CVF International Conference on Computer Vision}, 9619--9628.

\bibitem[{Kumar et~al.(2022)Kumar, Raghunathan, Jones, Ma, and Liang}]{kumar2022finetuning}
Kumar, A.; Raghunathan, A.; Jones, R.~M.; Ma, T.; and Liang, P. 2022.
\newblock Fine-Tuning can Distort Pretrained Features and Underperform Out-of-Distribution.
\newblock In \emph{International Conference on Learning Representations}.

\bibitem[{Levine et~al.(2016)Levine, Finn, Darrell, and Abbeel}]{levine2016end}
Levine, S.; Finn, C.; Darrell, T.; and Abbeel, P. 2016.
\newblock End-to-end training of deep visuomotor policies.
\newblock \emph{The Journal of Machine Learning Research}, 17(1): 1334--1373.

\bibitem[{Li et~al.(2018{\natexlab{a}})Li, Yang, Song, and Hospedales}]{li2018learning}
Li, D.; Yang, Y.; Song, Y.-Z.; and Hospedales, T. 2018{\natexlab{a}}.
\newblock Learning to generalize: Meta-learning for domain generalization.
\newblock In \emph{Proceedings of the AAAI conference on artificial intelligence}, volume~32.

\bibitem[{Li et~al.(2017)Li, Yang, Song, and Hospedales}]{cite:ICCV17DBA}
Li, D.; Yang, Y.; Song, Y.-Z.; and Hospedales, T.~M. 2017.
\newblock Deeper, broader and artier domain generalization.
\newblock In \emph{IEEE International Conference on Computer Vision (ICCV)}, 5543--5551. IEEE.

\bibitem[{Li et~al.(2018{\natexlab{b}})Li, Pan, Wang, and Kot}]{li2018domain}
Li, H.; Pan, S.~J.; Wang, S.; and Kot, A.~C. 2018{\natexlab{b}}.
\newblock Domain generalization with adversarial feature learning.
\newblock In \emph{Proceedings of the IEEE conference on computer vision and pattern recognition}, 5400--5409.

\bibitem[{Li et~al.(2018{\natexlab{c}})Li, Tian, Gong, Liu, Liu, Zhang, and Tao}]{li2018deep}
Li, Y.; Tian, X.; Gong, M.; Liu, Y.; Liu, T.; Zhang, K.; and Tao, D. 2018{\natexlab{c}}.
\newblock Deep domain generalization via conditional invariant adversarial networks.
\newblock In \emph{Proceedings of the European Conference on Computer Vision (ECCV)}, 624--639.

\bibitem[{Li et~al.(2019)Li, Yang, Zhou, and Hospedales}]{li2019feature}
Li, Y.; Yang, Y.; Zhou, W.; and Hospedales, T.~M. 2019.
\newblock Feature-Critic Networks for Heterogeneous Domain Generalization.
\newblock In \emph{International Conference on Machine Learning (ICML)}, volume~97, 3915--3924.

\bibitem[{Liang, Li, and Srikant(2017)}]{liang2017enhancing}
Liang, S.; Li, Y.; and Srikant, R. 2017.
\newblock Enhancing the reliability of out-of-distribution image detection in neural networks.
\newblock \emph{arXiv preprint arXiv:1706.02690}.

\bibitem[{Mancini et~al.(2020)Mancini, Akata, Ricci, and Caputo}]{cite:ECCV20Cumix}
Mancini, M.; Akata, Z.; Ricci, E.; and Caputo, B. 2020.
\newblock Towards Recognizing Unseen Categories in Unseen Domains.
\newblock In \emph{European Conference on Computer Vision (ECCV)}.

\bibitem[{Muandet, Balduzzi, and Sch{\"o}lkopf(2013)}]{muandet2013domain}
Muandet, K.; Balduzzi, D.; and Sch{\"o}lkopf, B. 2013.
\newblock Domain generalization via invariant feature representation.
\newblock In \emph{International Conference on Machine Learning}, 10--18. PMLR.

\bibitem[{Paszke et~al.(2019)Paszke, Gross, Massa, Lerer, Bradbury, Chanan, Killeen, Lin, Gimelshein, Antiga, Desmaison, Kopf, Yang, DeVito, Raison, Tejani, Chilamkurthy, Steiner, Fang, Bai, and Chintala}]{NEURIPS2019_9015}
Paszke, A.; Gross, S.; Massa, F.; Lerer, A.; Bradbury, J.; Chanan, G.; Killeen, T.; Lin, Z.; Gimelshein, N.; Antiga, L.; Desmaison, A.; Kopf, A.; Yang, E.; DeVito, Z.; Raison, M.; Tejani, A.; Chilamkurthy, S.; Steiner, B.; Fang, L.; Bai, J.; and Chintala, S. 2019.
\newblock PyTorch: An Imperative Style, High-Performance Deep Learning Library.
\newblock In \emph{Advances in Neural Information Processing Systems 32}, 8024--8035. Curran Associates, Inc.

\bibitem[{Peng et~al.(2019)Peng, Bai, Xia, Huang, Saenko, and Wang}]{peng2019moment}
Peng, X.; Bai, Q.; Xia, X.; Huang, Z.; Saenko, K.; and Wang, B. 2019.
\newblock Moment matching for multi-source domain adaptation.
\newblock In \emph{IEEE International Conference on Computer Vision (ICCV)}, 1406--1415.

\bibitem[{Peng et~al.(2017)Peng, Usman, Kaushik, Hoffman, Wang, and Saenko}]{visda2017}
Peng, X.; Usman, B.; Kaushik, N.; Hoffman, J.; Wang, D.; and Saenko, K. 2017.
\newblock VisDA: The Visual Domain Adaptation Challenge.

\bibitem[{Piratla, Netrapalli, and Sarawagi(2020)}]{piratla2020efficient}
Piratla, V.; Netrapalli, P.; and Sarawagi, S. 2020.
\newblock Efficient domain generalization via common-specific low-rank decomposition.
\newblock In \emph{International Conference on Machine Learning}, 7728--7738. PMLR.

\bibitem[{Saenko et~al.(2010)Saenko, Kulis, Fritz, and Darrell}]{cite:ECCV10Office}
Saenko, K.; Kulis, B.; Fritz, M.; and Darrell, T. 2010.
\newblock Adapting Visual Category Models to New Domains.
\newblock In \emph{European Conference on Computer Vision (ECCV)}.

\bibitem[{Saito et~al.(2019)Saito, Kim, Sclaroff, Darrell, and Saenko}]{saito2019semi}
Saito, K.; Kim, D.; Sclaroff, S.; Darrell, T.; and Saenko, K. 2019.
\newblock Semi-supervised domain adaptation via minimax entropy.
\newblock In \emph{Proceedings of the IEEE/CVF international conference on computer vision}, 8050--8058.

\bibitem[{Shankar et~al.(2018)Shankar, Piratla, Chakrabarti, Chaudhuri, Jyothi, and Sarawagi}]{shankar2018generalizing}
Shankar, S.; Piratla, V.; Chakrabarti, S.; Chaudhuri, S.; Jyothi, P.; and Sarawagi, S. 2018.
\newblock Generalizing Across Domains via Cross-Gradient Training.
\newblock In \emph{International Conference on Learning Representations}.

\bibitem[{Shu et~al.(2021)Shu, Cao, Wang, Wang, and Long}]{shu2021open}
Shu, Y.; Cao, Z.; Wang, C.; Wang, J.; and Long, M. 2021.
\newblock Open domain generalization with domain-augmented meta-learning.
\newblock In \emph{Proceedings of the IEEE/CVF Conference on Computer Vision and Pattern Recognition}, 9624--9633.

\bibitem[{Vaze et~al.(2022)Vaze, Han, Vedaldi, and Zisserman}]{vaze2022openset}
Vaze, S.; Han, K.; Vedaldi, A.; and Zisserman, A. 2022.
\newblock Open-Set Recognition: A Good Closed-Set Classifier is All You Need.
\newblock In \emph{International Conference on Learning Representations}.

\bibitem[{Venkateswara et~al.(2017)Venkateswara, Eusebio, Chakraborty, and Panchanathan}]{cite:CVPR17OfficeHome}
Venkateswara, H.; Eusebio, J.; Chakraborty, S.; and Panchanathan, S. 2017.
\newblock Deep Hashing Network for Unsupervised Domain Adaptation.
\newblock In \emph{IEEE Conference on Computer Vision and Pattern Recognition (CVPR)}.

\bibitem[{Wang et~al.(2019)Wang, Ge, Lipton, and Xing}]{cite:NIPS19PAR}
Wang, H.; Ge, S.; Lipton, Z.; and Xing, E.~P. 2019.
\newblock Learning robust global representations by penalizing local predictive power.
\newblock In \emph{Advances in Neural Information Processing Systems (NeurIPS)}, 10506--10518.

\bibitem[{Xie, Ma, and Liang(2021)}]{xie2021composed}
Xie, S.~M.; Ma, T.; and Liang, P. 2021.
\newblock Composed fine-tuning: Freezing pre-trained denoising autoencoders for improved generalization.
\newblock In \emph{International Conference on Machine Learning}, 11424--11435. PMLR.

\bibitem[{You et~al.(2019)You, Long, Cao, Wang, and Jordan}]{you2019universal}
You, K.; Long, M.; Cao, Z.; Wang, J.; and Jordan, M.~I. 2019.
\newblock Universal domain adaptation.
\newblock In \emph{Proceedings of the IEEE/CVF conference on computer vision and pattern recognition}, 2720--2729.

\bibitem[{Zhai et~al.(2022)Zhai, Wang, Mustafa, Steiner, Keysers, Kolesnikov, and Beyer}]{zhai2022lit}
Zhai, X.; Wang, X.; Mustafa, B.; Steiner, A.; Keysers, D.; Kolesnikov, A.; and Beyer, L. 2022.
\newblock Lit: Zero-shot transfer with locked-image text tuning.
\newblock In \emph{Proceedings of the IEEE/CVF Conference on Computer Vision and Pattern Recognition}, 18123--18133.

\bibitem[{Zhao et~al.(2020)Zhao, Gong, Liu, Fu, and Tao}]{zhao2020domain}
Zhao, S.; Gong, M.; Liu, T.; Fu, H.; and Tao, D. 2020.
\newblock Domain generalization via entropy regularization.
\newblock \emph{Advances in Neural Information Processing Systems}, 33: 16096--16107.

\bibitem[{Zhou, Ye, and Zhan(2021)}]{zhou2021learning}
Zhou, D.-W.; Ye, H.-J.; and Zhan, D.-C. 2021.
\newblock Learning placeholders for open-set recognition.
\newblock In \emph{Proceedings of the IEEE/CVF Conference on Computer Vision and Pattern Recognition}, 4401--4410.

\bibitem[{Zhou et~al.(2020)Zhou, Yang, Hospedales, and Xiang}]{zhou2020learning}
Zhou, K.; Yang, Y.; Hospedales, T.; and Xiang, T. 2020.
\newblock Learning to generate novel domains for domain generalization.
\newblock In \emph{European conference on computer vision}, 561--578. Springer.

\bibitem[{Zhou et~al.(2021)Zhou, Yang, Qiao, and Xiang}]{zhou2021domain}
Zhou, K.; Yang, Y.; Qiao, Y.; and Xiang, T. 2021.
\newblock Domain Generalization with MixStyle.
\newblock In \emph{International Conference on Learning Representations}.

\bibitem[{Zhu and Li(2021)}]{zhu2021crossmatch}
Zhu, R.; and Li, S. 2021.
\newblock CrossMatch: Cross-Classifier Consistency Regularization for Open-Set Single Domain Generalization.
\newblock In \emph{International Conference on Learning Representations}.

\end{thebibliography}

\end{document}